\documentclass[logo, address]{trfr}

\usepackage{natbib}

\usepackage{xcolor}

\usepackage{graphicx}

\usepackage{algorithm,algpseudocode}
\usepackage{algorithmicx}

\usepackage{multirow}

\usepackage{amssymb}

\usepackage{threeparttable}

\usepackage{hyperref}

\usepackage[most]{tcolorbox}
\usepackage{listings}

\usepackage{soul}
\definecolor{my_lightblue}{rgb}{0.85,0.95,1.0}
\sethlcolor{my_lightblue}

\title{Knowledge Graph-Assisted LLM Post-Training for Enhanced Legal Reasoning}

\abstract{
LLM post-training has primarily relied on large text corpora and human feedback, without capturing the structure of domain knowledge. This has caused models to struggle dealing with complex reasoning tasks, especially for high-stakes professional domains. In Law, reasoning requires deep understanding of the relations between various legal concepts, a key component missing in current LLM post-training. In this paper, we propose a knowledge graph (KG)-assisted approach for enhancing LLMs' reasoning capability in Legal that is generalizable to other high-stakes domains. We model key legal concepts by following the \textbf{IRAC} (Issue, Rule, Analysis and Conclusion) framework, and construct a KG with 12K legal cases. We then produce training data using our IRAC KG, and conduct both Supervised Fine-Tuning (SFT) and Direct Preference Optimization (DPO) with three state-of-the-art (SOTA) LLMs (30B, 49B and 70B), varying architecture and base model family. Our post-trained models obtained better average performance on 4/5 diverse legal benchmarks (14 tasks) than baselines. In particular, our 70B DPO model achieved the best score on 4/6 reasoning tasks, among baselines and a 141B SOTA legal LLM, demonstrating the effectiveness of our KG for enhancing LLMs' legal reasoning capability.
}

\author[1]{Dezhao Song}
\author[1]{Guglielmo Bonifazi}
\author[1,*]{Frank Schilder}
\author[1,2,*]{Jonathan Richard Schwarz}

\affiliation[1]{Thomson Reuters Foundational Research}
\affiliation[2]{Imperial College London}

\contribution[*]{Equal contribution}

\correspondence{\{first.last\}@thomsonreuters.com}

\begin{document}
\maketitle

\section{Introduction}
 
Large language models (LLMs), such as GPT \cite{DBLP:journals/corr/abs-2303-08774}, Claude \cite{anthropic2023claude} and Llama \cite{DBLP:journals/corr/abs-2307-09288}, have shown remarkable performance on a wide range of natural language tasks. However, when the target applications are in highly specialized domains, such as Medicine, Finance, Law, etc., general-purpose LLMs often struggle with specialized terminologies and domain‑specific reasoning needs. Consequently, researchers have begun to build domain‑specific variants by specifically pre‑training and fine-tuning on corpora from the corresponding target fields \cite{singhal2023large,Singhal:2025aa,xie-etal-2024-efficient,DBLP:conf/nips/ColomboPBMHMCCD24}.


While training and continuous fine-tuning on raw text capture the generation of domain language, they are insufficient to encode the structure of domain knowledge, especially on domain-specific reasoning tasks. Recent reasoning models work best for areas, such as mathematics and programming, where verifiable rewards enable a straightforward approach to reinforcement learning; however, areas like Medicine or Law require alternative forms of reasoning, such as argumentative reasoning or inference in the presence of high and inherent uncertainty, by using concepts of the corresponding domains, such as symptoms, events, and legal rules.

More importantly, these areas require deep understanding of the relations between abstract concepts. Financial market-event analysis adopts a cause-effect-impact-conclusion chain; clinical reasoning follows a symptom-diagnosis-treatment-outcome flow; in Legal, judicial reasoning employs the fact-rule-conclusion process. Encoding these reasoning patterns in a structured form offers a principled way to augment LLMs' inductive bias and to reduce hallucinations or misattributions.

In this paper, we focus on the legal domain where even when models are trained on vast corpora of case law, statutes, regulations, and secondary literature, they may produce misattributions, hallucinations, and logically incoherent arguments \cite{doi.org/10.1111/jels.12413}, and specific tasks are often better served by smaller specialized models \cite{DBLP:journals/corr/abs-2304-12202,jayakumar-etal-2023-large,DBLP:conf/cikm/ShuZLDDZ24}. The root of this problem lies not in the quantity of text for LLM training but in the structure of legal knowledge: legal prose is dense, highly contextual, and governed by rules that change over time, calling for a clearer textual representation that makes the element of the reasoning process more accessible.

In order to address this gap, we start with the formal modeling of legal concepts, which then leads to generating textual training data for LLM post-training. Our approach explicitly encodes the reasoning patterns found in judicial opinions in order to facilitate LLM training. The cornerstone of legal reasoning is the \textbf{IRAC} framework, \textit{\textbf{I}ssue}, \textit{\textbf{R}ule}, \textit{\textbf{A}nalysis} and \textit{\textbf{C}onclusion}, which is taught in law schools and encapsulates the typical flow of legal argumentation \cite{brand1976legalwriting}. By constructing an \textbf{IRAC Knowledge Graph} (IRAC KG) that captures the relationships between issues, governing rules, factual applications, and conclusions, we aim to provide a structured substrate from which an LLM can learn to reason more faithfully. 

Our work equips LLMs with a structured substrate for legal reasoning, but with two distinct features. First, instead of focusing on creating legal KGs with specific entities, such as Person, Organization, etc., our goal is to model the key legal concepts by following the IRAC framework. Second, rather than treating the graph merely as retrieval scaffolding, we post-train LLMs directly on graph-derived supervision signals, and show that this yields measurable gains on diverse legal tasks. Our contributions are threefold:
\begin{enumerate}
    \item \textbf{IRAC KG construction.} We extract IRAC components from a large corpus of 12K legal case opinions, and encode them as nodes and typed edges in a graph, preserving the causal dependencies that underlie legal reasoning.
    
    \item \textbf{IRAC KG-based LLM post-training.} By utilizing the KG as a knowledge source, we post-train mid to large LLMs (Qwen3-30B-A3B-Instruct-2507, Llama-3\_3-Nemotron-Super-49B-v1\_5 and Llama-3.1-70B-Instruct) via both Supervised Fine-Tuning (SFT) and Direct Preference Optimization (DPO) \cite{DBLP:conf/nips/RafailovSMMEF23} in order to improve their ability to generate coherent legal arguments.
    
    \item \textbf{Extensive evaluation.} We assess the trained models on five legal benchmarks: LexGLUE \cite{DBLP:journals/corr/abs-2304-12202}, LegalBench \cite{DBLP:conf/nips/GuhaNHRCKCPWRZT23}, COLIEE \cite{DBLP:journals/rss/RabeloGKKYS22}, SuperGPQA (Law) \cite{DBLP:journals/corr/abs-2502-14739} and $\delta$-Stance \cite{DBLP:conf/acl/GuptaR025}, demonstrating measurable gains over baseline and SOTA models.
\end{enumerate}


\section{Related Work}
\label{section:related_work}

\subsection{LLM-based KG Construction}
\label{section:related_work_kg_construction}

KG creation pipelines typically consist of two steps: Named Entity Recognition (NER) and Relation Extraction (RE). Prompting-based strategies have proven to be effective, addressing both tasks jointly \cite{DBLP:journals/corr/abs-2404-16130,guo2024lightrag} or separately \cite{jimenez2024hipporag,gutierrez2025from}. Compared to traditional methods, LLMs are also capable of extracting additional metadata, such as verbose entity descriptions or numerical scores measuring relation strength \cite{guo2024lightrag}.

Focusing on KG quality, \citet{zhang2024extract} defined a three-stage solution to ground LLMs into existing KGs, or to canonicalize one derived via open extraction. \citet{DBLP:conf/acl/HanCBS24} trained small models for output quality verification and iteratively improving the quality of the original KG, while \citet{zhu2024llms} developed a multi-agent framework to guide this refinement process by integrating external knowledge.

\subsection{Graph-based LLM Reasoning}
\label{section:related_work_llm_reasoning}

Researchers have utilized LLMs and graphs to address complex reasoning scenarios \cite{pan2024unifying}. GraphRAG \cite{DBLP:journals/corr/abs-2306-16092} organizes text corpora as a KG, and reasons over community descriptions to answer queries. This enables LLMs to handle questions that require a general understanding of the corpus and inspired follow-up methods to explore structured representations to improve Retrieval Augmented Generation (RAG) pipelines \cite{peng2024graph, han2024retrieval}.

A recurring theme in several of these works is to present the retrieved information to LLMs in structured formats, e.g., triples or relational paths, which has improved reasoning fidelity and quality \cite{DBLP:conf/iclr/MaXJLQYMG25,sun2024thinkongraph,liu2024knowledge,wen2024mindmap}. Rather than using graphs for retrieving structured information, \citet{cheng2024structure} reformulate the input query as a graph, and propose a multi-stage prompting pipeline to reason over the input graph for query answering. The authors claimed that the method reflects how humans often approach complex problems.

In addition to fostering reasoning, graphs were also used to model the reasoning process itself. Both Tree-of-Thoughts \cite{yao2023tree} and Graph-of-Thoughts \cite{besta2024got} improved over the classical Chain-of-Thought idea that lacks the capability of expressing complex thought processes, such as backtracking on unpromising steps or combining ideas from different branches.

\subsection{KG-LLM Synergy in the Legal Domain}
\label{section:kg_llm_legal}

\citet{tang2024caselink} represented legal cases and charges as nodes, and connected them using existing metadata, such as case-case citations and case-charge relationships. \citet{dhani2021similar} developed a more complex pipeline to extract entities and relations from Indian cases, judgments and laws using traditional NER and RE techniques for the purpose of finding similar cases. 

More recently, \citet{chen2024leverage} used an LLM to extract triples from Chinese cases and criminal law articles by following a pre-defined schema.
Similar to our approach, \citet{tang2024casegnn++,tang2024casegnn} relied on an LLM to represent case legal issues and facts as a KG (though without considering precedents). This offered an initial, though partial, representation of a case via the IRAC framework.

While being effective in solving important tasks, \citet{HANNAH2025100843} showed that serious legal implications might arise from LLM responses, and proposed a KG-RAG solution to address potential legal concerns. Similarly, \citet{10.1145/3627673.3680268} combined KG and RAG to develop an Italian legislative platform to support law analysis, drafting, landscape identification and complexity monitoring. Finally, ChatLaw \cite{DBLP:journals/corr/abs-2306-16092} is a multi-agent system for legal consulting, and utilizes KGs in intermediate steps to assist answering user requests.

Instead of extracting specific entities (e.g., Person, Location, etc.), our work builds KGs that model key legal concepts (such as Facts and Rules). Also, different from existing methods that combine LLMs with graphs, our novelty lies in the use of graph structure for domain-specific LLM post-training. Finally, while previous approaches studied partial modeling of legal reasoning, our IRAC KG offers a more comprehensive coverage of the important legal reasoning components.

\section{System Overview and KG Design}
\label{section:overview_and_kg_design}

\subsection{Overview}
\label{section:overview}

\begin{figure}[ht]
  \center
  \includegraphics[width=0.5\columnwidth]{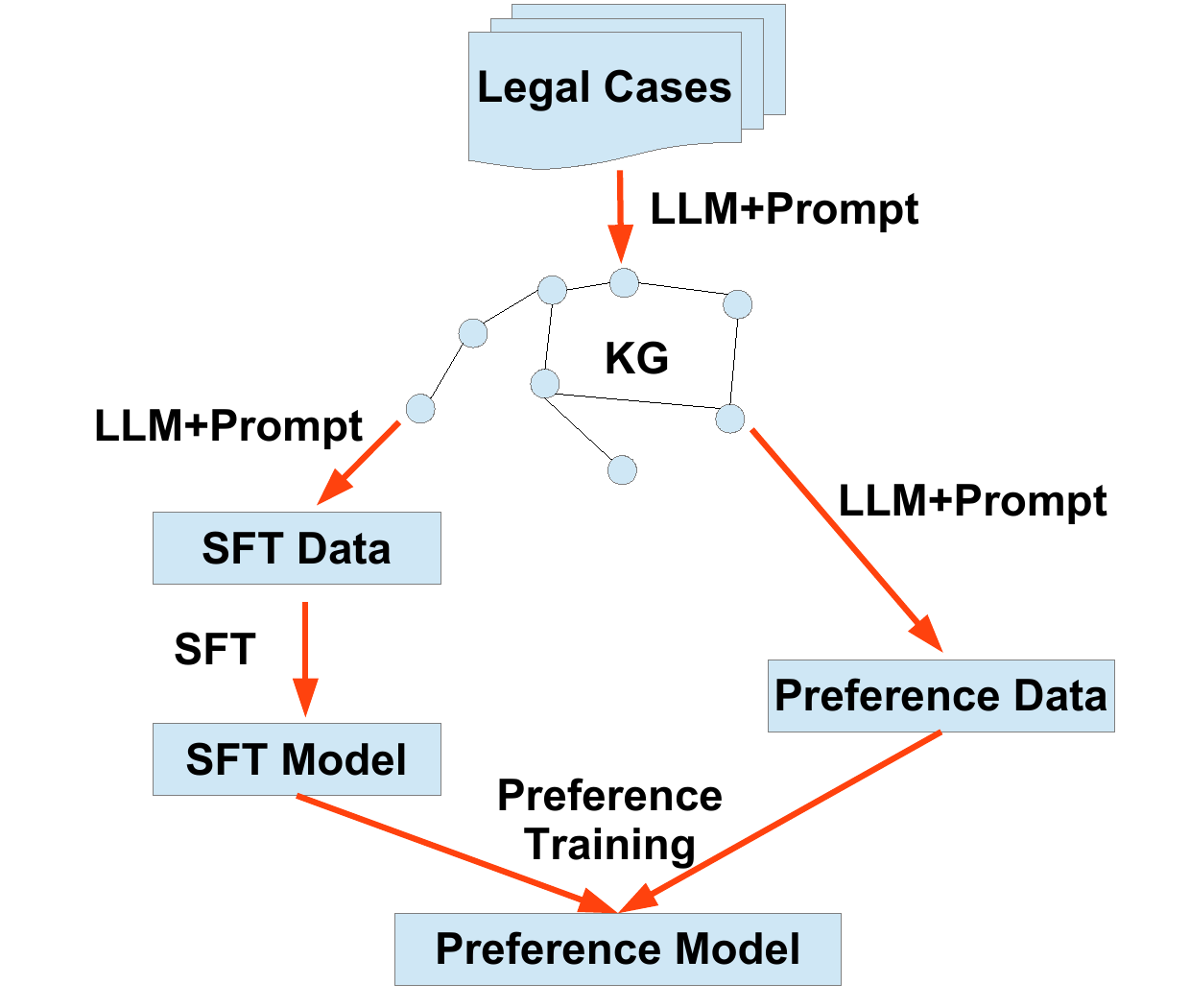}
  \caption{System overview.}
  \label{figure:overview}
\end{figure}

Figure \ref{figure:overview} demonstrates an overview of our entire knowledge graph-assisted LLM post-training system. The first step is to extract the key concepts from legal cases (from U.S. jurisdictions) by utilizing an LLM and a prompt (Appendix \ref{section:appendix_prompt_kg}) that includes the KG schema (Section \ref{section:kg_design}). Then, we produce training data for both Supervised Fine-Tuning (SFT) and Preference Training. During data generation, in addition to prompts, we also employed an LLM for explanation generation (Section \ref{section:rules_sft}) and as an LLM Judge (Section \ref{section:rules_dpo}).

\subsection{Knowledge Graph Design}
\label{section:kg_design}

\begin{figure}[ht]
  \center
  \includegraphics[width=0.5\columnwidth]{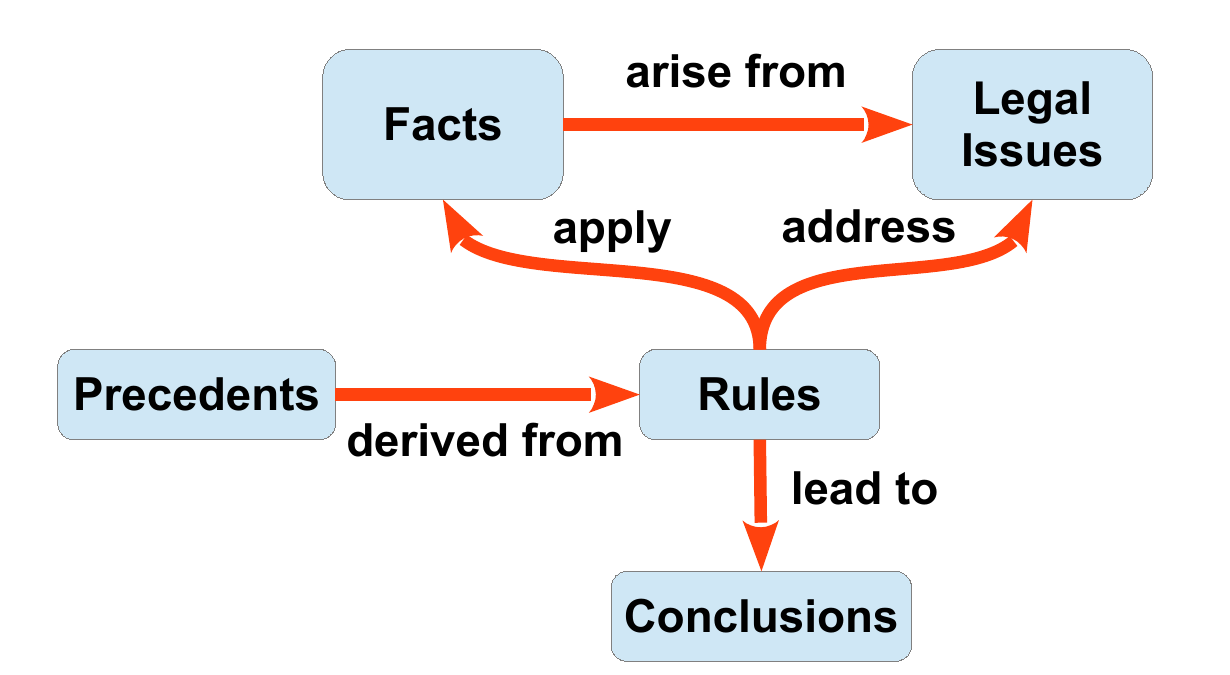}
  \caption{Schema of our IRAC legal knowledge graph.}
  \label{figure:kg_schema}
\end{figure}

Different from prior works that extract specific entity names (e.g., Person, Company, Location, etc.), our KG focuses on extracting the key legal concepts (such as Facts and Legal Issues) in order to facilitate legal reasoning. Figure \ref{figure:kg_schema} shows the design of our legal KG, a key component of our proposed system, by following the IRAC framework.

\begin{figure*}[t]
  \center
  \includegraphics[width=1.0\linewidth]{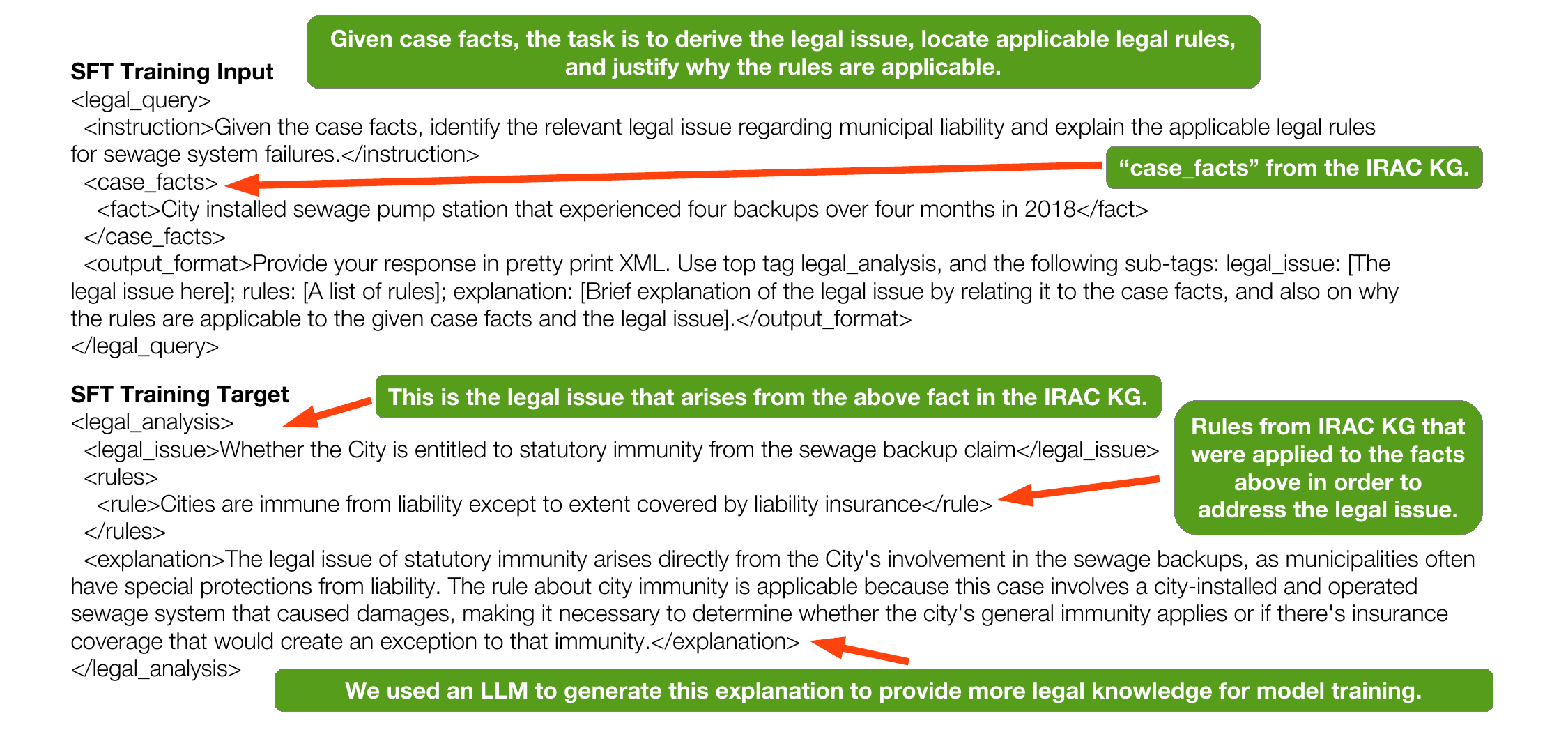}
  \caption{An example of the Rules SFT dataset.}
  \label{figure:rules_sft}
\end{figure*}

IRAC is commonly adopted for conducting legal analysis \cite{Burton2017ThinkLikeALawyer}. Even though there are variations, such as ILAC (Issue, Law, Analysis, Conclusion), IRAACP (Issue, Rule, Apply, Apply, Conclusion, Policy), etc., the core idea of legal reasoning centers around the same main concepts, i.e., Issue, Rule, Analysis and Conclusion. We consulted four legal experts on the design in order to capture the core of the legal reasoning process. Our IRAC KG includes the following components:
\begin{itemize}
  \item Legal Issues (\textbf{I}): This represents legal issues that the different parties (e.g., plaintiffs, defendants, judges, attorneys, etc.) are addressing in a legal case. Facts are things that actually happened, and the reason we have a legal case is due to the facts. Therefore, we have \textit{Legal Issues} \textbf{arise from} \textit{Facts} in our schema.
  \item Rules (\textbf{R}): This refers to legal rules. In the legal domain, prior cases (also known as precedents) that are sufficiently representative become rules that courts will reference to decide on future cases. Therefore, \textit{Rules} are \textbf{derived from} \textit{Precedents}.
  \item Analysis or Application (\textbf{A}): This denotes the legal reasoning process. When deciding on a legal case, courts will \textbf{apply} \textit{Rules} to \textit{Facts} in order to \textbf{address} the \textit{Legal Issues}.
  \item Conclusions (\textbf{C}): The analysis that courts carry out will eventually \textbf{lead to} certain \textit{Conclusions}. These are courts' rulings on the case.
\end{itemize}

\section{Training Data Generation}
\label{section:data_generation}

Our goal is to study whether such legal reasoning KGs would benefit LLMs on diverse tasks by conducting both SFT and Preference Optimization. In this section, we present how we utilized our IRAC KG to produce the corresponding training data.

\begin{figure*}[ht]
  \center
  \includegraphics[width=1.0\linewidth]{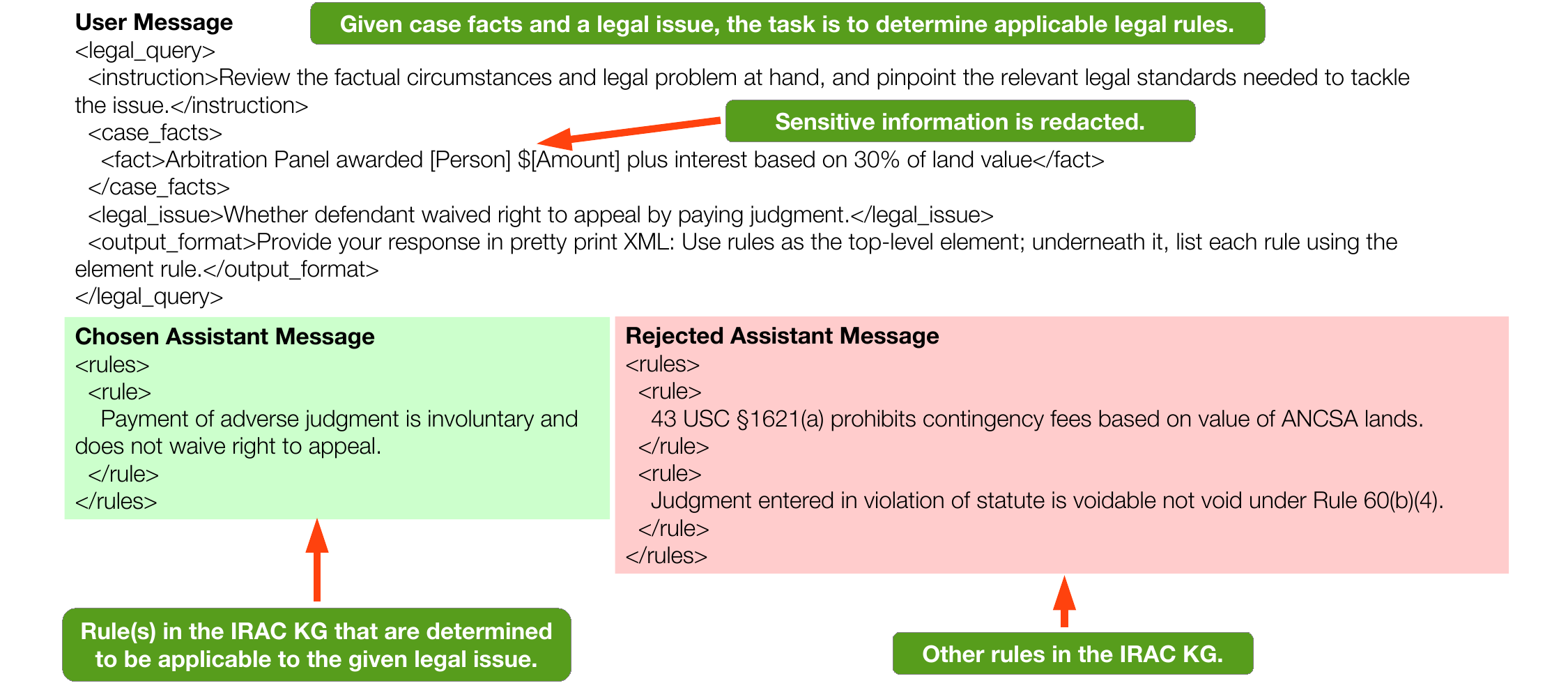}
  \caption{An example of the Rules preference dataset.}
  \label{figure:rules_dpo}
\end{figure*}

\subsection{The Rules SFT Dataset}
\label{section:rules_sft}

Figure \ref{figure:rules_sft} illustrates a concrete example of our SFT dataset. The core idea is to simulate how courts identify rule(s) that are applicable to addressing a specific legal issue. In SFT input, the main component is \textit{case\_facts}, which is the foundation to have a legal case. Then, in the training target, we teach a model how to derive the legal issue that arises from the fact(s), and further identify the applicable rule(s). In addition, we also provide an explanation on why the rule(s) are applicable to the legal issue.

Algorithm \ref{algorithm:rules_sft} depicts this SFT data generation process.
\begin{algorithm}[ht]
\caption{GenSFT($KG, l_{i}, LLM$), $KG$ represents the IRAC KG; $l_{i}$ denotes a legal issue in $KG$; a large language model $LLM$ produces an explanation on the applicability of certain rule(s) to $l_{i}$.}
\label{algorithm:rules_sft}

\begin{algorithmic}[1]
\State{$Rules \gets \emptyset$}
\State{$Facts \gets GetRelatedFacts(KG, l_{i})$}
\ForAll {$f \in Facts$}
	\State{$R_{f} \gets GetRsViaApply(KG, f)$}
	\State{$Rules \gets Rules \cup R_{f}$}
\EndFor
\State{$OtherRs \gets GetRsViaAddress(KG, l_{i})$}
\State{$Rules \gets Rules \cup OtherRs$}
\State{$Exp \gets Explain(LLM, l_{i}, Rules, Facts)$}
\State{\Return {$Facts, Rules, Exp$}}
\end{algorithmic}
\end{algorithm}
At Line 2, $Facts$ consists of all facts related to legal issue $l_{i}$, via the \textit{\textbf{arise from}} relation in Figure \ref{figure:kg_schema}. Next, from Line 3 to 8, we find the set of applicable rules to $l_{i}$ both directly and indirectly. On one hand, at Line 4, for each fact $f$ related to $l_{i}$, Function $GetRsViaApply(KG, f)$ finds the rules that were applied to $f$ by following the path: \textit{Rules$ \rightarrow$ \textbf{apply} $\rightarrow$ Facts}. On the other hand, at Line 7, by following the \textit{Rules $\rightarrow$ \textbf{address} $\rightarrow$ Legal Issues} chain in Figure \ref{figure:kg_schema}, Function $GetRsViaAddress(KG, l_{i})$ gathers additional rules identified by the court that could be used to address $l_{i}$ but may not actually be applied to any facts, e.g., rules from similar precedents.
As demonstrated in Figure \ref{figure:rules_sft}, the retrieved facts become SFT input, while the rules and the legal issue itself are part of the training target.

In addition, at Line 9, we prompt (Appendix \ref{section:appendix_prompt_sft}) an LLM to produce a concise explanation on why the rules are applicable to the legal issue $l_{i}$, hoping to provide more legal knowledge to downstream training tasks; this explanation constitutes another section of the training target. We conducted a quality review of 15 SFT examples with three subject matter experts (SMEs), where each example was labeled by a single SME. Among these, 11, 4 and 0 examples were graded as \textit{Correct}, \textit{Correct with minor issues} and \textit{Wrong} respectively, verifying the quality of our SFT dataset.



\begin{table*}[ht]
  \centering
  \begin{tabular}{l|cccc|cccc}
    \hline
   \multirow{2}{*}{\textbf{Entity}} & \multicolumn{4}{c|}{\textbf{Claude Sonnet 3.5 KG}} & \multicolumn{4}{c}{\textbf{Llama-3.1-405B-Instruct KG}} \\
    & \textbf{Good} & \textbf{Acceptable} & \textbf{Poor} & \textbf{M} & \textbf{Good} & \textbf{Acceptable} & \textbf{Poor} & \textbf{M} \\
    \hline
    Fact & 68\% & 5\% & 2\% & 26\% & 50\% & 9\% & 6\% & 35\% \\ \hline
    Legal Issue & 63\% & 0\% & 0\% & 37\% & 75\% & 5\% & 10\% & 10\% \\ \hline
    Rule & 69\% & 0\% & 4\% & 27\% & 29\% & 0\%& 33\% & 38\% \\ \hline
    Conclusion & 68\% & 0\% & 0\% & 32\% & 60\% & 0\% & 10\% & 30\% \\ \hline \hline
    \multirow{2}{*}{\textbf{Relation}} & \multicolumn{2}{c}{\textbf{Pass}} & \multicolumn{2}{c|}{\textbf{Fail}} & \multicolumn{2}{c}{\textbf{Pass}} & \multicolumn{2}{c}{\textbf{Fail}} \\
    & \multicolumn{2}{c}{92\%} & \multicolumn{2}{c|}{8\%} & \multicolumn{2}{c}{59\%} & \multicolumn{2}{c}{41\%} \\ \hline
  \end{tabular}
  \caption{\label{table:kg_quality_review}
    IRAC KG quality review. \textit{Fact}, \textit{Legal Issue}, \textit{Rule} and \textit{Conclusion} correspond to the entity types in our IRAC KG in Figure \ref{figure:kg_schema}; we report a single number for all relations. For entities, our SMEs labeled them as \textbf{\textit{Good}} (high-quality with potentially minor inaccuracies), \textbf{\textit{Acceptable}} (with noticeable issues but still useful) and \textbf{\textit{Poor}} (with significant errors or completely incorrect); our SMEs also checked whether the KG missed any entities (\textbf{\textit{M}}) by reading the entire case. As for relations, our review only examines whether they are correct or not. Note that if any of the entities involved in a relation is judged as \textbf{\textit{Poor}}, then this relation will be labeled as \textbf{\textit{Fail}}. We determined these quality review criteria together with our domain experts.
  }
\end{table*}

\subsection{The Rules Preference Dataset}
\label{section:rules_dpo}

In addition to SFT, we also performed Preference Optimization in this work. Similar to the SFT dataset, we produced our preference data by utilizing the IRAC KG with a focus on teaching the models to correctly identify the proper legal rules to cope with a given legal issue. Figure \ref{figure:rules_dpo} shows a concrete example of this dataset. Different from the above SFT data, our preference training input (i.e., \textit{User Message}) includes both \textit{case\_facts} and \textit{legal\_issue}, with chosen and rejected legal rules as the training target. Our hypothesis is that by teaching models how to select the appropriate rules for different legal matters, they would be able to gain more knowledge on how courts reason in a variety of legal situations.

We formally present the data generation process in Algorithm \ref{algorithm:rules_dpo}.
\begin{algorithm}[ht]
\caption{GenPref($KG, l_{i}, LJ$), $KG$ denotes the IRAC KG; $l_{i}$ is a legal issue in $KG$; an LLM Judge $LJ$ determines a rule's applicability to $l_{i}$.}
\label{algorithm:rules_dpo}

\begin{algorithmic}[1]
\State{$Facts \gets GetRelatedFacts(KG, l_{i})$}
\State{$Chosen\_Rs \gets \emptyset$}
\ForAll {$f \in Facts$}
	\State{$R_{f} \gets GetRsViaApply(KG, f)$}
	\State{$Chosen\_Rs \gets Chosen\_Rs \cup R_{f}$}
\EndFor
\State{$OtherRs \gets GetRsViaAddress(KG, l_{i})$}
\State{$Chosen\_Rs \gets Chosen\_Rs \cup OtherRs$}
\State $Rejected\_Rs \gets All\_Rs \setminus Chosen\_Rs$
\ForAll {$r \in Rejected\_Rs$}
	\If {$Applicable(LJ, Facts, l_{i}, r)$}
		\State {$Rejected\_Rs \gets Rejected\_Rs \setminus \{r\}$}
	\EndIf
\EndFor
\State {\Return {$Facts, Chosen\_Rs, Rejected\_Rs$}}
\end{algorithmic}
\end{algorithm}
For preference training, we need to provide both the chosen and the rejected rules. Similar to SFT data generation, at Line 1, we first obtain facts ($Facts$) related to the legal issue $l_{i}$. Then, from Line 2 to 8, we find the chosen rules ($Chosen\_Rs$) that are either directly (via the \textit{\textbf{address}} relation) or indirectly (via the combination of the \textit{\textbf{apply}} and the \textit{\textbf{arise from}} relations) connected to $l_{i}$. In order to produce the rejected rules, from Line 9 to 14, we use an LLM judge (Appendix \ref{section:appendix_prompt_dpo}) to determine whether each of the non-chosen rules is truly not applicable. The intuition is that even though a rule is not directly or indirectly connected to $l_{i}$ in the KG, it may be due to an imperfect KG generation process. The output of the LLM Judge could be: \textit{Yes}/\textit{No}/\textit{Potentially}, and we only keep the rules that are determined as non-applicable (i.e., the LLM Judge outputs \textit{No}).

\subsection{Model Training}
\label{section:model_training}

In our work, we conduct both SFT and Preference Optimization, DPO \cite{DBLP:conf/nips/RafailovSMMEF23} specifically. By giving models the case facts, SFT (Figure \ref{figure:rules_sft}) teaches the models how to identify the legal issue, locate the applicable rules, and learn the rationale behind these materials (\textit{Exp} at Line 9 in Algorithm \ref{algorithm:rules_sft}). Different from SFT, our preference data (Figure \ref{figure:rules_dpo}) focuses on teaching the models how to conduct legal reasoning by contrasting the applicable and non-applicable legal rules with respect to a variety of legal matters. For DPO, we adopted the default Loss function with a small NLL term on the chosen answers \cite{DBLP:journals/corr/abs-2404-19733}.

We emphasize that the above training data is not directly related to our evaluation benchmarks (Section \ref{section:evaluation_benchmarks}). Our primary goal is to provide the models with high-quality domain knowledge via different training paradigms in order to study whether the trained models would be able to gain improved capability on diverse tasks in the same domain.

\section{Evaluation}
\label{section:evaluation}

\subsection{IRAC KG Generation and Quality Review}
\label{section:evaluation_kg_generation}

We randomly sampled $\sim$12K cases, with 220 cases from each U.S. jurisdiction, and divided them into Train and Validation for model training purposes with a ratio of 10:1. We produced KGs with both Claude Sonnet 3.5 and Llama-3.1-405B-Instruct. Two SMEs then reviewed the Sonnet KG of 18 cases and the Llama KG of 8 cases\footnote{Due to its inferior quality, we stopped the review process.}, with each entity and relation examined by a single SME.

Table \ref{table:kg_quality_review} shows that the Sonnet KG has superior quality on most entity types (except for \textit{Legal Issue}), by taking into account both \textit{Good} and \textit{Acceptable}. Note that the Llama KG generally has notably higher ratios of entities labeled as \textit{Poor}. In terms of relations, the Sonnet KG also showed much higher quality, with only 8\% of relations labeled as \textit{Fail} compared to that of 41\% of the Llama KG.

One drawback of both KGs is that they miss a decent percentage of entities. In a legal case, not all facts or issues are considered critical for courts to decide their rulings. A legal issue may be mentioned in passing without further discussions; also, some facts may merely be included in a case while not being essential to the core issues of the case. In such situations, missing entities have little impact in understanding the case. We leave it to future work to improve our KG's comprehensiveness.

Given the above quality review, we adopted the IRAC KG from Claude Sonnet 3.5 for producing the post-training datasets. We also utilized Sonnet 3.5 to generate the explanation ($Exp$ at Line 9 in Algorithm \ref{algorithm:rules_sft}) for the SFT dataset, and employed it as the LLM Judge ($LJ$ at Line 11 in Algorithm \ref{algorithm:rules_dpo}) during preference data generation.

\subsection{Evaluation Benchmarks}
\label{section:evaluation_benchmarks}

In order to verify the effectiveness of our proposed IRAC KG, we carried out extensive experiments on diverse legal datasets (detailed in Appendix \ref{section:appendix_benchmark}).

\textbf{LexGLUE}\footnote{https://github.com/coastalcph/zeroshot\_lexglue} \cite{DBLP:journals/corr/abs-2304-12202} is a well-adopted legal benchmark, covering European and U.S. jurisdictions. It mainly consists of (multi-class and multi-label) classification tasks, together with a multiple choice question answering task; each task includes 1,000 samples. The number of classes range from 10 to 100, presenting substantial challenges to legal AI systems.

\textbf{LegalBench} \cite{DBLP:conf/nips/GuhaNHRCKCPWRZT23} is a recent benchmark that mainly focuses on multiple choice question answering. In addition, it also includes several information extraction tasks, such as extracting plaintiffs and defendants from case excerpts. It includes a total of 162 tasks, and we removed \textit{rule\_qa} whose evaluation requires manual review according to the LegalBench team.

\textbf{COLIEE} \cite{DBLP:journals/rss/RabeloGKKYS22} is a well-known competition for legal systems, with two major types of tasks: retrieval (of relevant cases or statues) and entailment. For retrieval, it is to find supporting prior cases to help decide on a new case in Canada (Task-1) or to locate relevant law articles (statutes) for Japanese legal queries (Task-3). For entailment, instead of requiring the systems to locate the relevant prior cases or statues, the input already includes this information, and the systems should decide whether the following entailments hold: <paragraph (from a prior legal case) $\models$ court decision> (Task-2) or <statute $\models$ legal\_query> (Task-4). Given that the test set is relatively small for each specific year, we combined the data from all previous years for our evaluation.

\textbf{SuperGPQA} \cite{DBLP:journals/corr/abs-2502-14739} consists of graduate-level multiple choice questions for a broad range of domains, e.g., Medicine, History, etc. Our evaluation focuses on the 656 questions in \textit{Law}, where each question contains 9 $\sim$ 10 choices.

\textbf{$\delta$-Stance} \cite{DBLP:conf/acl/GuptaR025} is a large-scale dataset to specifically evaluate AI systems' legal reasoning capability. In U.S. legal cases, introductory signals (e.g., \textit{see, contra, see generally, etc.}) between two text snippets often indicate certain important relations. This dataset consists of $\sim$14K triples of <snippet A, signal, snippet B>. We consulted SMEs to group such triples into four higher-level classes (Appendix \ref{section:delta_stance}): directly supports, indirectly supports, contradicts and background. We also produced a 3-class version where we removed the ``background'' category that does not provide as much value as the other three according to SMEs.


\subsection{Model Training Details}
\label{section:model_training_details}

We post-trained (Appendix \ref{section:hyperparameter}) models of different scales: \textbf{Qwen3-30B-A3B-Instruct-2507} (Qwen3-30B), \textbf{
Llama-3\_3-Nemotron-Super-49B-v1\_5} (Nemotron1.5-49B) and \textbf{Llama-3.1-70B-Instruct} (Llama3.1-70B). The first two recent models obtained competitive results on various benchmarks. As for Llama \cite{DBLP:journals/corr/abs-2407-21783}, the 3.3 model is not as competitive as the 3.1 model (Appendix \ref{section:llama}), thus we utilized the earlier version.


\subsection{Results and Discussion}
\label{section:results}

\begin{table*}[ht]
  \centering
  \begin{threeparttable}
  \begin{tabular}{cccc|ccc|ccc|c}
    \hline
    \multirow{2}{*}{\textbf{Benchmark}} & \multicolumn{3}{c|}{\textbf{Qwen3-30B}} & \multicolumn{3}{c|}{\textbf{Nemotron1.5-49B}} & \multicolumn{3}{c|}{\textbf{Llama3.1-70B}} & \textbf{SaulLM} \\
    & OOB & SFT & DPO & OOB & SFT & DPO & OOB & SFT & DPO & (141B) \\ \hline
    \multicolumn{1}{l}{LexGLUE Average} & 57.9 & 60.3 & \textbf{61.6} & 57.0 & 61.2 & \textbf{62.5} & 65.0 & 66.8 & \hl{\textbf{67.2}} & 64.4 \\
    ECtHR A\tnote{$\bigstar$} & 53.4 & 59.7 & \underline{62.8} & 52.8 & 63.2 & \underline{66.3} & 73.6 & 77.3 & \underline{\hl{77.9}} & 73.6 \\
    ECtHR B\tnote{$\bigstar$} & 60.2 & 66.3 & \underline{69.0} & 60.9 & 70.0 & \underline{73.7} & 80.0 & 82.1 & \underline{\hl{82.3}} & 77.6 \\
    EUR-LEX\tnote{$\bigstar$} & 27.6 & 28.5 & \underline{30.5} & 27.1 & 29.2 & \underline{29.5} & \underline{\hl{33.9}} & 32.3 & 31.8 & 32.2 \\
    LEDGAR\tnote{$\bigstar$} & \underline{69.7} & 69.6 & 69.6 & 66.0 & 66.5 & \underline{67.2} & 63.6 & 66.8 & \underline{67.5} & \hl{71.8} \\
    SCOTUS\tnote{$\bigstar$} & \underline{66.0} & 65.8 & 65.9 & 65.4 & \underline{66.0} & 65.4 & 67.5 & \underline{\hl{68.9}} & 68.8 & 52.2 \\
    CaseHOLD\tnote{$\S$$\Diamond$} & 70.7 & 71.7 & \underline{72.0} & 69.7 & 72.5 & \underline{73.2} & 71.6 & 73.4 & \underline{75.1} & \hl{79.0} \\
   \hline
    \multicolumn{1}{l}{COLIEE Average} & 56.6 & 59.0 & \textbf{61.1} & 49.4 & 53.5 & \textbf{54.2} & 57.0 & 59.8 & \hl{\textbf{60.7}} & 24.6 \\
    Task-1\tnote{$\bigstar$} & 39.8 & 35.0 & \underline{\hl{40.3}} & 20.3 & 28.3 & \underline{28.5} & 30.8 & 30.6 & \underline{32.4} & 0.0 \\
    Task-2\tnote{$\bigstar$$\Diamond$} & 52.2 & 60.0 & \underline{62.1} & 48.0 & 54.6 & \underline{57.2} & 57.7 & 66.0 & \underline{\hl{67.6}} & 25.1 \\
    Task-3\tnote{$\triangle$} & 64.3 & 67.1 & \underline{\hl{67.7}} & \underline{53.6} & 53.4 & 53.4 & 61.3 & 64.4 & \underline{64.6} & 0.0 \\
    Task-4\tnote{$\S$$\Diamond$} & 70.2 & 73.9 & \underline{74.2} & 75.9 & 77.5 & \underline{77.7} & \underline{\hl{78.4}} & 77.9 & 78.0 & 73.1 \\
   \hline
   \multicolumn{1}{l}{$\delta$-Stance Average\tnote{$\dag$$\Diamond$}} & \textbf{45.5} & 44.3 & 43.7 & 33.9 & 38.7 & \textbf{40.9} & 48.8 & 49.7 & \hl{\textbf{50.1}} & 43.8 \\
   3-class & \underline{52.7} & 51.2 & 50.7 & 38.8 & 44.0 & \underline{46.5} & 55.1 & 56.1 & \underline{\hl{56.5}} & 48.8 \\
   4-class & \underline{38.3} & 37.3 & 36.8 & 29.1 & 33.4 & \underline{35.3} & 42.6 & 43.2 & \underline{\hl{43.6}} & 38.8 \\
   \hline
   \multicolumn{1}{l}{LegalBench\tnote{$\blacktriangle$}} & 74.1 & 75.3 & \textbf{75.4} & 74.2 & 75.9 & \textbf{76.4} & 78.0 & \hl{\textbf{79.3}} & \hl{\textbf{79.3}} & \hl{79.3} \\
   \hline
   \multicolumn{1}{l}{SuperGPQA (Law)\tnote{$\S$$\Diamond$}} & 42.4 & 42.1 & \textbf{42.5} & 37.7 & 37.6 & \textbf{38.0} & 40.3 & 41.7 & \hl{\textbf{43.8}} & 28.4 \\
   \hline
  \end{tabular}
  \begin{tablenotes}
    \item $\Diamond$: Legal reasoning task.
    \item $\bigstar$: Micro-F1; $\dag$: Macro-F1; $\triangle$: Macro-F2; $\S$: Accuracy; $\blacktriangle$: Balanced Accuracy.
  \end{tablenotes}
  \end{threeparttable}
  \caption{\label{table:results}
    Evaluation results (0$\sim$100). OOB denotes the out-of-the-box model without additional training; SFT and DPO represent the models after the corresponding training respectively. For each model, we underscore each task's best performance, and bold the highest average performance; blue background indicates the highest score for a row.
  }
\end{table*}

\textbf{SFT/DPO vs. OOB}. In Table \ref{table:results}, compared to the OOB models, our KG-based post-training achieved improvements on 10, 13 and 12 out of the 14 tasks for Qwen3-30B, Nemotron1.5-49B and Llama3.1-70B respectively, where the best performance falls on either SFT or DPO. Furthermore, for all three models, DPO training helped obtain the highest average performance on most benchmarks, except for Qwen3-30B on $\delta$-Stance.

\textbf{DPO vs. SFT}. Compared to SFT, DPO did provide additional benefits on 11/14 tasks for Qwen3-30B and Llama3.1-70B, and on 12/14 tasks for Nemotron1.5-49B. Unlike prior works that mostly focus on training smaller models, such as 7B/8B \cite{DBLP:conf/aaai/FanHW0JS25,DBLP:journals/corr/abs-2507-00018,DBLP:journals/corr/abs-2404-14723} and 32B \cite{wen-etal-2025-light}, our results demonstrated the benefits of KG-based DPO training of different and also larger scales: 30B, 49B and 70B. Furthermore, compared to the 30B model, the two larger DPO models exhibited clearer advantages than SFT on 4/6 legal reasoning tasks: CaseHOLD, $\delta$-Stance (both tasks) and SuperGPQA.

\textbf{SaulLM (141B)} \cite{DBLP:conf/nips/ColomboPBMHMCCD24} is a SOTA legal LLM that was trained on a large amount of legal data. Despite post-trained on much smaller SFT (13.5K) and DPO (5.3K) datasets, our models show noticeable benefits. First, while Nemotron1.5-49B OOB is substantially worse than SaulLM on LexGLUE, $\delta$-Stance and LegalBench, its DPO version clearly narrowed the gaps. Furthermore, although Llama3.1-70B OOB already outperforms SaulLM on most benchmarks, our post-training did fully close the gap on LegalBench.


\textbf{Instruction Following (IF)}. Since Nemotron1.5-49B is also based on Llama, we only present the scores of Qwen3-30B and Llama3.1-70B in Table \ref{table:if}. In general, both models were able to follow the instructions in our prompts. Furthermore, our post-training did not damage models' IF capability, and in fact, they provided noticeable improvements: Qwen3-30B on LexGLUE, and Llama3.1-70B on COLIEE and SuperGPQA (Law).
\begin{table}[ht]
  \centering
  \begin{tabular}{lcc}
    \hline
    \multirow{2}{*}{\textbf{Benchmark}} & \textbf{Qwen3-30B} & \textbf{Llama3.1-70B} \\
    & O/S/D & O/S/D \\ \hline
    LexGLUE & 5.7/3.4/2.1 & 1.8/2.7/2.2 \\
    COLIEE & 0.4/0.3/0.2 & 3.4/1.8/2.1 \\
    $\delta$-Stance & 0.0 & 0.0 \\
    LegalBench & 0.4/0.2/0.2 & 0.2/0.1/0.2 \\
    SuperGPQA & 0.0 & 1.4/0.0/0.2 \\ \hline
  \end{tabular}
  \caption{\label{table:if}
    IF error rates (0$\sim$100). We report the average of all tasks for each benchmark. O, S and D represent the OOB, SFT and DPO models respectively; a single number indicates the same rate for all models.
  }
\end{table}

We did not employ complex post-processings: We lowercased model outputs, removed punctuations, and performed exact string matching against the groundtruth. Due to our relatively strict metric calculation, compared to other benchmarks, we do observe higher error rates on LexGLUE (mainly classification tasks), where we require the predictions of a test sample to be a subset of the available labels for the corresponding task. For instance, for EUR-LEX (multi-label classification of EU legislative documents), Qwen3-30B OOB predicated \textit{risk management} while only \textit{management} is a valid label. We do not give partial credit in such scenarios.


\section{Conclusion}
\label{section:conclusion}

In this paper, we proposed a generalizable knowledge graph-assisted approach for enhancing LLMs' legal reasoning capability. We follow the IRAC framework in order to model courts' analysis and reasoning process of legal cases. Furthermore, we conducted both SFT and DPO training with mid to large models (30B, 49B and 70B) by utilizing data generated from the IRAC KG. Our trained models achieved better average performance on 4/5 diverse legal benchmarks than baselines. Compared to SFT, the Llama3.1-70B DPO model exhibited promising results on 3/6 legal reasoning tasks, with 1.7\% (CaseHOLD), 1.6\% (COLIEE-Task-2) and 2.1\% (SuperGPQA Law) higher scores respectively; in particular, it achieved noticeably higher scores than SaulLM, a 141B SOTA legal LLM trained on a large amount of legal data.

\section*{Limitations}
\label{section:limitation}

Our current work is limited to U.S. Law (common law), and the generalizability of our approach to other legal systems warrants further exploration. In addition to legal, the generalizability of our proposed KG-assisted model post-training for other high-stakes domains would also be an important topic. Furthermore, in this work, we produced our data with 12K U.S. cases. It would be interesting to utilize more cases, and also study the impact of additional relations, e.g., citations and overrulings; however, we need to be mindful of the environmental impact, since it would require more GPU hours. Also, given the emergence of other LLMs, we should explore alternatives other than Claude Sonnet 3.5 for KG and data creation, and compare the performance of the resulting models. Finally, although our models showed advantages on diverse legal benchmarks, it would be important to deploy them in front of users to gather feedback on real-world tasks, especially those that require complex legal reasoning. When deploying the trained models for real-world use, it would be ideal to apply certain guardrails to prevent potential malicious usage, especially for the legal domain.


\bibliographystyle{plainnat}
\bibliography{kg}

\begin{thebibliography}{51}
\providecommand{\natexlab}[1]{#1}
\providecommand{\url}[1]{\texttt{#1}}
\expandafter\ifx\csname urlstyle\endcsname\relax
  \providecommand{\doi}[1]{doi: #1}\else
  \providecommand{\doi}{doi: \begingroup \urlstyle{rm}\Url}\fi

\bibitem[Anthropic(2023)]{anthropic2023claude}
Anthropic.
\newblock Introducing {C}laude.
\newblock \url{https://www.anthropic.com/index/claude}, 2023.

\bibitem[Besta et~al.(2024)Besta, Blach, Kubicek, Gerstenberger, Gianinazzi,
  Gajda, Lehmann, Podstawski, Niewiadomski, Nyczyk, and Hoefler]{besta2024got}
Maciej Besta, Nils Blach, Ales Kubicek, Robert Gerstenberger, Lukas Gianinazzi,
  Joanna Gajda, Tomasz Lehmann, Micha{\l} Podstawski, Hubert Niewiadomski,
  Piotr Nyczyk, and Torsten Hoefler.
\newblock {Graph of Thoughts: Solving Elaborate Problems with Large Language
  Models}.
\newblock \emph{Proceedings of the AAAI Conference on Artificial Intelligence},
  38\penalty0 (16):\penalty0 17682--17690, Mar 2024.
\newblock \doi{10.1609/aaai.v38i16.29720}.
\newblock URL \url{https://ojs.aaai.org/index.php/AAAI/article/view/29720}.

\bibitem[Brand and White(1976)]{brand1976legalwriting}
Norman Brand and John~O. White.
\newblock \emph{Legal Writing: The Strategy of Persuasion}.
\newblock St. Martin's Press, New York, 1976.

\bibitem[Burton(2017)]{Burton2017ThinkLikeALawyer}
Kelley Burton.
\newblock ``think like a lawyer'' using a legal reasoning grid and
  criterion-referenced assessment rubric on irac (issue, rule, application,
  conclusion).
\newblock \emph{Journal of Learning Design}, 10\penalty0 (2):\penalty0 57--66,
  2017.
\newblock \doi{10.5204/jld.v10i2.229}.
\newblock URL \url{https://www.jld.edu.au/article/view/229.html}.

\bibitem[Chalkidis(2023)]{DBLP:journals/corr/abs-2304-12202}
Ilias Chalkidis.
\newblock Chat{GPT} may pass the bar exam soon, but has a long way to go for
  the {LexGLUE} benchmark.
\newblock \emph{CoRR}, abs/2304.12202, 2023.
\newblock \doi{10.48550/ARXIV.2304.12202}.
\newblock URL \url{https://doi.org/10.48550/arXiv.2304.12202}.

\bibitem[Chen et~al.(2024)Chen, Chen, Zhu, Pei, Chen, Zhou, Wang, Zhou, Li, and
  Zhang]{chen2024leverage}
Yongming Chen, Miner Chen, Ye~Zhu, Juan Pei, Siyu Chen, Yu~Zhou, Yi~Wang, Yifan
  Zhou, Hao Li, and Songan Zhang.
\newblock Leverage knowledge graph and large language model for law article
  recommendation: A case study of chinese criminal law.
\newblock \emph{arXiv preprint arXiv:2410.04949}, 2024.

\bibitem[Cheng et~al.(2024)Cheng, Ahmed, Willke, and Sun]{cheng2024structure}
Kewei Cheng, Nesreen Ahmed, Theodore Willke, and Yizhou Sun.
\newblock Structure guided prompt: Instructing large language model in
  multi-step reasoning by exploring graph structure of the text.
\newblock In \emph{Proceedings of the 2024 Conference on Empirical Methods in
  Natural Language Processing}, pages 9407--9430, 2024.

\bibitem[Colombo(2024)]{10.1145/3627673.3680268}
Andrea Colombo.
\newblock Leveraging knowledge graphs and llms to support and monitor
  legislative systems.
\newblock In \emph{Proceedings of the 33rd ACM International Conference on
  Information and Knowledge Management}, CIKM '24, pages 5443--5446, New York,
  NY, USA, 2024. Association for Computing Machinery.
\newblock ISBN 9798400704369.
\newblock \doi{10.1145/3627673.3680268}.
\newblock URL \url{https://doi.org/10.1145/3627673.3680268}.

\bibitem[Colombo et~al.(2024)Colombo, Pires, Boudiaf, Melo, Hautreux,
  Malaboeuf, Charpentier, Culver, and Desa]{DBLP:conf/nips/ColomboPBMHMCCD24}
Pierre Colombo, Telmo~Pessoa Pires, Malik Boudiaf, Rui Melo, Gabriel Hautreux,
  Etienne Malaboeuf, Johanne Charpentier, Dominic Culver, and Michael Desa.
\newblock Saul{LM}-54{B} {\&} {S}aul{LM}-141{B}: Scaling up domain adaptation
  for the legal domain.
\newblock In Amir Globersons, Lester Mackey, Danielle Belgrave, Angela Fan,
  Ulrich Paquet, Jakub~M. Tomczak, and Cheng Zhang, editors, \emph{Advances in
  Neural Information Processing Systems 38: Annual Conference on Neural
  Information Processing Systems 2024, NeurIPS 2024, Vancouver, BC, Canada,
  December 10 - 15, 2024}, 2024.
\newblock URL
  \url{http://papers.nips.cc/paper\_files/paper/2024/hash/ea3f85a33f9ba072058e3df233cf6cca-Abstract-Conference.html}.

\bibitem[Cui et~al.(2023)Cui, Li, Yan, Chen, and
  Yuan]{DBLP:journals/corr/abs-2306-16092}
Jiaxi Cui, Zongjian Li, Yang Yan, Bohua Chen, and Li~Yuan.
\newblock {ChatLaw}: Open-source legal large language model with integrated
  external knowledge bases.
\newblock \emph{CoRR}, abs/2306.16092, 2023.
\newblock \doi{10.48550/ARXIV.2306.16092}.
\newblock URL \url{https://doi.org/10.48550/arXiv.2306.16092}.

\bibitem[Dhani et~al.(2021)Dhani, Bhatt, Ganesan, Sirohi, and
  Bhatnagar]{dhani2021similar}
Jaspreet~Singh Dhani, Ruchika Bhatt, Balaji Ganesan, Parikshet Sirohi, and
  Vasudha Bhatnagar.
\newblock Similar cases recommendation using legal knowledge graphs.
\newblock In \emph{ACM SIGKDD International Conference on Knowledge Discovery
  and Data Mining}, 2021.

\bibitem[Dubey et~al.(2024)Dubey, Jauhri, Pandey, Kadian, Al{-}Dahle, Letman,
  Mathur, Schelten, Yang, Fan, Goyal, Hartshorn, Yang, Mitra, Sravankumar,
  Korenev, Hinsvark, Rao, Zhang, Rodriguez, Gregerson, Spataru, Rozi{\`{e}}re,
  Biron, Tang, Chern, Caucheteux, Nayak, Bi, Marra, McConnell, Keller, Touret,
  Wu, Wong, Ferrer, Nikolaidis, Allonsius, Song, Pintz, Livshits, Esiobu,
  Choudhary, Mahajan, Garcia{-}Olano, Perino, Hupkes, Lakomkin, AlBadawy,
  Lobanova, Dinan, Smith, Radenovic, Zhang, Synnaeve, Lee, Anderson, Nail,
  Mialon, Pang, Cucurell, Nguyen, Korevaar, Xu, Touvron, Zarov, Ibarra,
  Kloumann, Misra, Evtimov, Copet, Lee, Geffert, Vranes, Park, Mahadeokar,
  Shah, van~der Linde, Billock, Hong, Lee, Fu, Chi, Huang, Liu, Wang, Yu,
  Bitton, Spisak, Park, Rocca, Johnstun, Saxe, Jia, Alwala, Upasani, Plawiak,
  Li, Heafield, Stone, and et~al.]{DBLP:journals/corr/abs-2407-21783}
Abhimanyu Dubey, Abhinav Jauhri, Abhinav Pandey, Abhishek Kadian, Ahmad
  Al{-}Dahle, Aiesha Letman, Akhil Mathur, Alan Schelten, Amy Yang, Angela Fan,
  Anirudh Goyal, Anthony Hartshorn, Aobo Yang, Archi Mitra, Archie Sravankumar,
  Artem Korenev, Arthur Hinsvark, Arun Rao, Aston Zhang, Aur{\'{e}}lien
  Rodriguez, Austen Gregerson, Ava Spataru, Baptiste Rozi{\`{e}}re, Bethany
  Biron, Binh Tang, Bobbie Chern, Charlotte Caucheteux, Chaya Nayak, Chloe Bi,
  Chris Marra, Chris McConnell, Christian Keller, Christophe Touret, Chunyang
  Wu, Corinne Wong, Cristian~Canton Ferrer, Cyrus Nikolaidis, Damien Allonsius,
  Daniel Song, Danielle Pintz, Danny Livshits, David Esiobu, Dhruv Choudhary,
  Dhruv Mahajan, Diego Garcia{-}Olano, Diego Perino, Dieuwke Hupkes, Egor
  Lakomkin, Ehab AlBadawy, Elina Lobanova, Emily Dinan, Eric~Michael Smith,
  Filip Radenovic, Frank Zhang, Gabriel Synnaeve, Gabrielle Lee, Georgia~Lewis
  Anderson, Graeme Nail, Gr{\'{e}}goire Mialon, Guan Pang, Guillem Cucurell,
  Hailey Nguyen, Hannah Korevaar, Hu~Xu, Hugo Touvron, Iliyan Zarov,
  Imanol~Arrieta Ibarra, Isabel~M. Kloumann, Ishan Misra, Ivan Evtimov, Jade
  Copet, Jaewon Lee, Jan Geffert, Jana Vranes, Jason Park, Jay Mahadeokar, Jeet
  Shah, Jelmer van~der Linde, Jennifer Billock, Jenny Hong, Jenya Lee, Jeremy
  Fu, Jianfeng Chi, Jianyu Huang, Jiawen Liu, Jie Wang, Jiecao Yu, Joanna
  Bitton, Joe Spisak, Jongsoo Park, Joseph Rocca, Joshua Johnstun, Joshua Saxe,
  Junteng Jia, Kalyan~Vasuden Alwala, Kartikeya Upasani, Kate Plawiak, Ke~Li,
  Kenneth Heafield, Kevin Stone, and et~al.
\newblock The {L}lama 3 herd of models.
\newblock \emph{CoRR}, abs/2407.21783, 2024.
\newblock \doi{10.48550/ARXIV.2407.21783}.
\newblock URL \url{https://doi.org/10.48550/arXiv.2407.21783}.

\bibitem[Edge et~al.(2024)Edge, Trinh, Cheng, Bradley, Chao, Mody, Truitt, and
  Larson]{DBLP:journals/corr/abs-2404-16130}
Darren Edge, Ha~Trinh, Newman Cheng, Joshua Bradley, Alex Chao, Apurva Mody,
  Steven Truitt, and Jonathan Larson.
\newblock From {L}ocal to {G}lobal: {A} {G}raph {RAG} approach to query-focused
  summarization.
\newblock \emph{CoRR}, abs/2404.16130, 2024.
\newblock \doi{10.48550/ARXIV.2404.16130}.
\newblock URL \url{https://doi.org/10.48550/arXiv.2404.16130}.

\bibitem[Fan et~al.(2025)Fan, Hong, Wang, Bao, Jiang, and
  Song]{DBLP:conf/aaai/FanHW0JS25}
Yuchen Fan, Yuzhong Hong, Qiushi Wang, Junwei Bao, Hongfei Jiang, and Yang
  Song.
\newblock Preference-oriented supervised fine-tuning: Favoring target model
  over aligned large language models.
\newblock In Toby Walsh, Julie Shah, and Zico Kolter, editors, \emph{AAAI-25,
  Sponsored by the Association for the Advancement of Artificial Intelligence,
  February 25 - March 4, 2025, Philadelphia, PA, {USA}}, pages 23859--23867.
  {AAAI} Press, 2025.
\newblock \doi{10.1609/AAAI.V39I22.34558}.
\newblock URL \url{https://doi.org/10.1609/aaai.v39i22.34558}.

\bibitem[Guha et~al.(2023)Guha, Nyarko, Ho, R{\'{e}}, Chilton, Aditya,
  Chohlas{-}Wood, Peters, Waldon, Rockmore, Zambrano, Talisman, Hoque, Surani,
  Fagan, Sarfaty, Dickinson, Porat, Hegland, Wu, Nudell, Niklaus, Nay, Choi,
  Tobia, Hagan, Ma, Livermore, Rasumov{-}Rahe, Holzenberger, Kolt, Henderson,
  Rehaag, Goel, Gao, Williams, Gandhi, Zur, Iyer, and
  Li]{DBLP:conf/nips/GuhaNHRCKCPWRZT23}
Neel Guha, Julian Nyarko, Daniel~E. Ho, Christopher R{\'{e}}, Adam Chilton,
  K.~Aditya, Alex Chohlas{-}Wood, Austin Peters, Brandon Waldon, Daniel~N.
  Rockmore, Diego Zambrano, Dmitry Talisman, Enam Hoque, Faiz Surani, Frank
  Fagan, Galit Sarfaty, Gregory~M. Dickinson, Haggai Porat, Jason Hegland,
  Jessica Wu, Joe Nudell, Joel Niklaus, John~J. Nay, Jonathan~H. Choi, Kevin
  Tobia, Margaret Hagan, Megan Ma, Michael~A. Livermore, Nikon Rasumov{-}Rahe,
  Nils Holzenberger, Noam Kolt, Peter Henderson, Sean Rehaag, Sharad Goel,
  Shang Gao, Spencer Williams, Sunny Gandhi, Tom Zur, Varun Iyer, and Zehua Li.
\newblock {LegalBench}: {A} collaboratively built benchmark for measuring legal
  reasoning in large language models.
\newblock In Alice Oh, Tristan Naumann, Amir Globerson, Kate Saenko, Moritz
  Hardt, and Sergey Levine, editors, \emph{Advances in Neural Information
  Processing Systems 36: Annual Conference on Neural Information Processing
  Systems 2023, NeurIPS 2023, New Orleans, LA, USA, December 10 - 16, 2023},
  2023.
\newblock URL
  \url{http://papers.nips.cc/paper\_files/paper/2023/hash/89e44582fd28ddfea1ea4dcb0ebbf4b0-Abstract-Datasets\_and\_Benchmarks.html}.

\bibitem[Guo et~al.(2024)Guo, Xia, Yu, Ao, and Huang]{guo2024lightrag}
Zirui Guo, Lianghao Xia, Yanhua Yu, Tu~Ao, and Chao Huang.
\newblock {LightRAG}: Simple and fast retrieval-augmented generation.
\newblock \emph{arXiv preprint arXiv:2410.05779}, 2024.

\bibitem[Gupta et~al.(2025)Gupta, Rice, and O'Connor]{DBLP:conf/acl/GuptaR025}
Ankita Gupta, Douglas Rice, and Brendan~T. O'Connor.
\newblock $\delta$-{S}tance: {A} large-scale real world dataset of stances in
  legal argumentation.
\newblock In Wanxiang Che, Joyce Nabende, Ekaterina Shutova, and Mohammad~Taher
  Pilehvar, editors, \emph{Proceedings of the 63rd Annual Meeting of the
  Association for Computational Linguistics (Volume 1: Long Papers), {ACL}
  2025, Vienna, Austria, July 27 - August 1, 2025}, pages 31450--31467.
  Association for Computational Linguistics, 2025.
\newblock URL \url{https://aclanthology.org/2025.acl-long.1517/}.

\bibitem[Guti{\'e}rrez et~al.(2025)Guti{\'e}rrez, Shu, Qi, Zhou, and
  Su]{gutierrez2025from}
Bernal~Jim{\'e}nez Guti{\'e}rrez, Yiheng Shu, Weijian Qi, Sizhe Zhou, and
  Yu~Su.
\newblock From {RAG} to memory: Non-parametric continual learning for large
  language models.
\newblock In \emph{Forty-second International Conference on Machine Learning},
  2025.
\newblock URL \url{https://openreview.net/forum?id=LWH8yn4HS2}.

\bibitem[Han et~al.(2024{\natexlab{a}})Han, Wang, Shomer, Guo, Ding, Lei,
  Halappanavar, Rossi, Mukherjee, Tang, et~al.]{han2024retrieval}
Haoyu Han, Yu~Wang, Harry Shomer, Kai Guo, Jiayuan Ding, Yongjia Lei, Mahantesh
  Halappanavar, Ryan~A Rossi, Subhabrata Mukherjee, Xianfeng Tang, et~al.
\newblock Retrieval-augmented generation with graphs (graphrag).
\newblock \emph{arXiv preprint arXiv:2501.00309}, 2024{\natexlab{a}}.

\bibitem[Han et~al.(2024{\natexlab{b}})Han, Collier, Buntine, and
  Shareghi]{DBLP:conf/acl/HanCBS24}
Jiuzhou Han, Nigel Collier, Wray~L. Buntine, and Ehsan Shareghi.
\newblock Pive: Prompting with iterative verification improving graph-based
  generative capability of llms.
\newblock In Lun{-}Wei Ku, Andre Martins, and Vivek Srikumar, editors,
  \emph{Findings of the Association for Computational Linguistics, {ACL} 2024,
  Bangkok, Thailand and virtual meeting, August 11-16, 2024}, pages 6702--6718.
  Association for Computational Linguistics, 2024{\natexlab{b}}.
\newblock \doi{10.18653/V1/2024.FINDINGS-ACL.400}.
\newblock URL \url{https://doi.org/10.18653/v1/2024.findings-acl.400}.

\bibitem[Hannah et~al.(2025)Hannah, Sousa, Dasoulas, and
  d'Amato]{HANNAH2025100843}
George Hannah, Rita~T. Sousa, Ioannis Dasoulas, and Claudia d'Amato.
\newblock On the legal implications of large language model answers: A prompt
  engineering approach and a view beyond by exploiting knowledge graphs.
\newblock \emph{Journal of Web Semantics}, 84:\penalty0 100843, 2025.
\newblock ISSN 1570-8268.
\newblock \doi{https://doi.org/10.1016/j.websem.2024.100843}.
\newblock URL
  \url{https://www.sciencedirect.com/science/article/pii/S1570826824000295}.

\bibitem[Hsu et~al.(2024)Hsu, Dai, Kothapalli, Song, Tang, Zhu, Shimizu, Sahni,
  Ning, and Chen]{DBLP:journals/corr/abs-2410-10989}
Pin{-}Lun Hsu, Yun Dai, Vignesh Kothapalli, Qingquan Song, Shao Tang, Siyu Zhu,
  Steven Shimizu, Shivam Sahni, Haowen Ning, and Yanning Chen.
\newblock Liger kernel: Efficient triton kernels for {LLM} training.
\newblock \emph{CoRR}, abs/2410.10989, 2024.
\newblock \doi{10.48550/ARXIV.2410.10989}.
\newblock URL \url{https://doi.org/10.48550/arXiv.2410.10989}.

\bibitem[Jayakumar et~al.(2023)Jayakumar, Farooqui, and
  Farooqui]{jayakumar-etal-2023-large}
Thanmay Jayakumar, Fauzan Farooqui, and Luqman Farooqui.
\newblock Large language models are legal but they are not: Making the case for
  a powerful {L}egal{LLM}.
\newblock In Daniel Preoțiuc-Pietro, Catalina Goanta, Ilias Chalkidis, Leslie
  Barrett, Gerasimos Spanakis, and Nikolaos Aletras, editors, \emph{Proceedings
  of the Natural Legal Language Processing Workshop 2023}, pages 223--229,
  Singapore, December 2023. Association for Computational Linguistics.
\newblock \doi{10.18653/v1/2023.nllp-1.22}.
\newblock URL \url{https://aclanthology.org/2023.nllp-1.22/}.

\bibitem[Jimenez~Gutierrez et~al.(2024)Jimenez~Gutierrez, Shu, Gu, Yasunaga,
  and Su]{jimenez2024hipporag}
Bernal Jimenez~Gutierrez, Yiheng Shu, Yu~Gu, Michihiro Yasunaga, and Yu~Su.
\newblock {HippoRAG}: Neurobiologically inspired long-term memory for large
  language models.
\newblock \emph{Advances in Neural Information Processing Systems},
  37:\penalty0 59532--59569, 2024.

\bibitem[Liu et~al.(2024)Liu, Wang, Zhu, Dong, and Li]{liu2024knowledge}
Haochen Liu, Song Wang, Yaochen Zhu, Yushun Dong, and Jundong Li.
\newblock Knowledge graph-enhanced large language models via path selection.
\newblock In \emph{Findings of the Association for Computational Linguistics
  ACL 2024}, pages 6311--6321, 2024.

\bibitem[Ma et~al.(2025)Ma, Xu, Jiang, Li, Qu, Yang, Mao, and
  Guo]{DBLP:conf/iclr/MaXJLQYMG25}
Shengjie Ma, Chengjin Xu, Xuhui Jiang, Muzhi Li, Huaren Qu, Cehao Yang, Jiaxin
  Mao, and Jian Guo.
\newblock Think-on-graph 2.0: Deep and faithful large language model reasoning
  with knowledge-guided retrieval augmented generation.
\newblock In \emph{The Thirteenth International Conference on Learning
  Representations, {ICLR} 2025, Singapore, April 24-28, 2025}. OpenReview.net,
  2025.
\newblock URL \url{https://openreview.net/forum?id=oFBu7qaZpS}.

\bibitem[Magesh et~al.(2025)Magesh, Surani, Dahl, Suzgun, Manning, and
  Ho]{doi.org/10.1111/jels.12413}
Varun Magesh, Faiz Surani, Matthew Dahl, Mirac Suzgun, Christopher~D. Manning,
  and Daniel~E. Ho.
\newblock {Hallucination-Free? Assessing the Reliability of Leading AI Legal
  Research Tools}.
\newblock \emph{Journal of Empirical Legal Studies}, 22\penalty0 (2):\penalty0
  216--242, 2025.
\newblock \doi{https://doi.org/10.1111/jels.12413}.
\newblock URL \url{https://onlinelibrary.wiley.com/doi/abs/10.1111/jels.12413}.

\bibitem[OpenAI(2023)]{DBLP:journals/corr/abs-2303-08774}
OpenAI.
\newblock {GPT-4} technical report.
\newblock \emph{CoRR}, abs/2303.08774, 2023.
\newblock \doi{10.48550/ARXIV.2303.08774}.
\newblock URL \url{https://doi.org/10.48550/arXiv.2303.08774}.

\bibitem[Pan et~al.(2024)Pan, Luo, Wang, Chen, Wang, and Wu]{pan2024unifying}
Shirui Pan, Linhao Luo, Yufei Wang, Chen Chen, Jiapu Wang, and Xindong Wu.
\newblock Unifying large language models and knowledge graphs: A roadmap.
\newblock \emph{IEEE Transactions on Knowledge and Data Engineering},
  36\penalty0 (7):\penalty0 3580--3599, 2024.

\bibitem[Pang et~al.(2024)Pang, Yuan, Cho, He, Sukhbaatar, and
  Weston]{DBLP:journals/corr/abs-2404-19733}
Richard~Yuanzhe Pang, Weizhe Yuan, Kyunghyun Cho, He~He, Sainbayar Sukhbaatar,
  and Jason Weston.
\newblock Iterative reasoning preference optimization.
\newblock \emph{CoRR}, abs/2404.19733, 2024.
\newblock \doi{10.48550/ARXIV.2404.19733}.
\newblock URL \url{https://doi.org/10.48550/arXiv.2404.19733}.

\bibitem[Peng et~al.(2024)Peng, Zhu, Liu, Bo, Shi, Hong, Zhang, and
  Tang]{peng2024graph}
Boci Peng, Yun Zhu, Yongchao Liu, Xiaohe Bo, Haizhou Shi, Chuntao Hong, Yan
  Zhang, and Siliang Tang.
\newblock Graph retrieval-augmented generation: A survey.
\newblock \emph{arXiv preprint arXiv:2408.08921}, 2024.

\bibitem[Rabelo et~al.(2022)Rabelo, Goebel, Kim, Kano, Yoshioka, and
  Satoh]{DBLP:journals/rss/RabeloGKKYS22}
Juliano Rabelo, Randy Goebel, Mi{-}Young Kim, Yoshinobu Kano, Masaharu
  Yoshioka, and Ken Satoh.
\newblock Overview and discussion of the competition on legal information
  extraction/entailment {(COLIEE)} 2021.
\newblock \emph{Rev. Socionetwork Strateg.}, 16\penalty0 (1):\penalty0
  111--133, 2022.
\newblock \doi{10.1007/S12626-022-00105-Z}.
\newblock URL \url{https://doi.org/10.1007/s12626-022-00105-z}.

\bibitem[Rafailov et~al.(2023)Rafailov, Sharma, Mitchell, Manning, Ermon, and
  Finn]{DBLP:conf/nips/RafailovSMMEF23}
Rafael Rafailov, Archit Sharma, Eric Mitchell, Christopher~D. Manning, Stefano
  Ermon, and Chelsea Finn.
\newblock Direct preference optimization: Your language model is secretly a
  reward model.
\newblock In Alice Oh, Tristan Naumann, Amir Globerson, Kate Saenko, Moritz
  Hardt, and Sergey Levine, editors, \emph{Advances in Neural Information
  Processing Systems 36: Annual Conference on Neural Information Processing
  Systems 2023, NeurIPS 2023, New Orleans, LA, USA, December 10 - 16, 2023},
  2023.
\newblock URL
  \url{http://papers.nips.cc/paper\_files/paper/2023/hash/a85b405ed65c6477a4fe8302b5e06ce7-Abstract-Conference.html}.

\bibitem[Saeidi et~al.(2024)Saeidi, Verma, and
  Baral]{DBLP:journals/corr/abs-2404-14723}
Amir Saeidi, Shivanshu Verma, and Chitta Baral.
\newblock Insights into alignment: Evaluating {DPO} and its variants across
  multiple tasks.
\newblock \emph{CoRR}, abs/2404.14723, 2024.
\newblock \doi{10.48550/ARXIV.2404.14723}.
\newblock URL \url{https://doi.org/10.48550/arXiv.2404.14723}.

\bibitem[Shu et~al.(2024)Shu, Zhao, Liu, Demeter, Du, and
  Zhang]{DBLP:conf/cikm/ShuZLDDZ24}
Dong Shu, Haoran Zhao, Xukun Liu, David Demeter, Mengnan Du, and Yongfeng
  Zhang.
\newblock {LawLLM}: Law large language model for the {US} legal system.
\newblock In Edoardo Serra and Francesca Spezzano, editors, \emph{Proceedings
  of the 33rd {ACM} International Conference on Information and Knowledge
  Management, {CIKM} 2024, Boise, ID, USA, October 21-25, 2024}, pages
  4882--4889. {ACM}, 2024.
\newblock \doi{10.1145/3627673.3680020}.
\newblock URL \url{https://doi.org/10.1145/3627673.3680020}.

\bibitem[Singhal et~al.(2023)Singhal, Azizi, Tu, Mahdavi, Wei, Chung, Scales,
  Tanwani, Cole-Lewis, Pfohl, et~al.]{singhal2023large}
Karan Singhal, Shekoofeh Azizi, Tao Tu, S~Sara Mahdavi, Jason Wei, Hyung~Won
  Chung, Nathan Scales, Ajay Tanwani, Heather Cole-Lewis, Stephen Pfohl, et~al.
\newblock Large language models encode clinical knowledge.
\newblock \emph{Nature}, 620\penalty0 (7972):\penalty0 172--180, 2023.

\bibitem[Singhal et~al.(2025)Singhal, Tu, Gottweis, Sayres, Wulczyn, Amin, Hou,
  Clark, Pfohl, Cole-Lewis, Neal, Rashid, Schaekermann, Wang, Dash, Chen, Shah,
  Lachgar, Mansfield, Prakash, Green, Dominowska, Ag{\"u}era~y Arcas, Toma{\v
  s}ev, Liu, Wong, Semturs, Mahdavi, Barral, Webster, Corrado, Matias, Azizi,
  Karthikesalingam, and Natarajan]{Singhal:2025aa}
Karan Singhal, Tao Tu, Juraj Gottweis, Rory Sayres, Ellery Wulczyn, Mohamed
  Amin, Le~Hou, Kevin Clark, Stephen~R. Pfohl, Heather Cole-Lewis, Darlene
  Neal, Qazi~Mamunur Rashid, Mike Schaekermann, Amy Wang, Dev Dash, Jonathan~H.
  Chen, Nigam~H. Shah, Sami Lachgar, Philip~Andrew Mansfield, Sushant Prakash,
  Bradley Green, Ewa Dominowska, Blaise Ag{\"u}era~y Arcas, Nenad Toma{\v s}ev,
  Yun Liu, Renee Wong, Christopher Semturs, S.~Sara Mahdavi, Joelle~K. Barral,
  Dale~R. Webster, Greg~S. Corrado, Yossi Matias, Shekoofeh Azizi, Alan
  Karthikesalingam, and Vivek Natarajan.
\newblock Toward expert-level medical question answering with large language
  models.
\newblock \emph{Nature Medicine}, 31\penalty0 (3):\penalty0 943--950, 2025.
\newblock \doi{10.1038/s41591-024-03423-7}.
\newblock URL \url{https://doi.org/10.1038/s41591-024-03423-7}.

\bibitem[Sun et~al.(2024)Sun, Xu, Tang, Wang, Lin, Gong, Ni, Shum, and
  Guo]{sun2024thinkongraph}
Jiashuo Sun, Chengjin Xu, Lumingyuan Tang, Saizhuo Wang, Chen Lin, Yeyun Gong,
  Lionel Ni, Heung-Yeung Shum, and Jian Guo.
\newblock {Think-on-Graph}: Deep and responsible reasoning of large language
  model on knowledge graph.
\newblock In \emph{The Twelfth International Conference on Learning
  Representations}, 2024.
\newblock URL \url{https://openreview.net/forum?id=nnVO1PvbTv}.

\bibitem[Tang et~al.(2024{\natexlab{a}})Tang, Qiu, Liu, Li, and
  Huang]{tang2024casegnn}
Yanran Tang, Ruihong Qiu, Yilun Liu, Xue Li, and Zi~Huang.
\newblock {CaseGNN}: Graph neural networks for legal case retrieval with
  text-attributed graphs.
\newblock In \emph{European conference on information retrieval}, pages 80--95.
  Springer, 2024{\natexlab{a}}.

\bibitem[Tang et~al.(2024{\natexlab{b}})Tang, Qiu, Liu, Li, and
  Huang]{tang2024casegnn++}
Yanran Tang, Ruihong Qiu, Yilun Liu, Xue Li, and Zi~Huang.
\newblock {CaseGNN}++: Graph contrastive learning for legal case retrieval with
  graph augmentation.
\newblock \emph{arXiv preprint arXiv:2405.11791}, 2024{\natexlab{b}}.

\bibitem[Tang et~al.(2024{\natexlab{c}})Tang, Qiu, Yin, Li, and
  Huang]{tang2024caselink}
Yanran Tang, Ruihong Qiu, Hongzhi Yin, Xue Li, and Zi~Huang.
\newblock {CaseLink}: Inductive graph learning for legal case retrieval.
\newblock In \emph{Proceedings of the 47th International ACM SIGIR Conference
  on Research and Development in Information Retrieval}, pages 2199--2209,
  2024{\natexlab{c}}.

\bibitem[Team et~al.(2025)Team, Du, Yao, Ma, Wang, Zheng, Zhu, Liu, Liang, Jin,
  Wei, Zheng, Deng, Jia, Jiang, Liao, Li, Li, Li, Li, Li, Ma, Ni, Que, Wang,
  Wen, Wu, Xing, Xu, Yang, Wang, Zhou, Bai, Bu, Cai, Chen, Chen, Cheng, Cheng,
  Ding, Huang, Huang, Li, Li, Li, Liang, Lin, Lin, Ma, Pang, Peng, Peng, Qi,
  Qiu, Qu, Quan, Tan, Wang, Wang, Wang, Wang, Wang, Xu, Yang, Yuan, Yue, Zhan,
  Zhang, Zhang, Zhang, Zhang, Zhang, Zhao, Zheng, Zhong, Gao, Li, Liu, Liu,
  Liu, Ni, Peng, Qin, Su, Wang, Wang, Yang, Yang, Cao, Yue, Zhang, Zhou, Liu,
  Lin, Huang, and Zhang]{DBLP:journals/corr/abs-2502-14739}
M.{-}A{-}P. Team, Xinrun Du, Yifan Yao, Kaijing Ma, Bingli Wang, Tianyu Zheng,
  Kang Zhu, Minghao Liu, Yiming Liang, Xiaolong Jin, Zhenlin Wei, Chujie Zheng,
  Kaixin Deng, Shian Jia, Sichao Jiang, Yiyan Liao, Rui Li, Qinrui Li, Sirun
  Li, Yizhi Li, Yunwen Li, Dehua Ma, Yuansheng Ni, Haoran Que, Qiyao Wang,
  Zhoufutu Wen, Siwei Wu, Tianshun Xing, Ming Xu, Zhenzhu Yang, Zekun~Moore
  Wang, Jun Zhou, Yuelin Bai, Xingyuan Bu, Chenglin Cai, Liang Chen, Yifan
  Chen, Chengtuo Cheng, Tianhao Cheng, Keyi Ding, Siming Huang, Yun Huang,
  Yaoru Li, Yizhe Li, Zhaoqun Li, Tianhao Liang, Chengdong Lin, Hongquan Lin,
  Yinghao Ma, Tianyang Pang, Zhongyuan Peng, Zifan Peng, Qige Qi, Shi Qiu,
  Xingwei Qu, Shanghaoran Quan, Yizhou Tan, Zili Wang, Chenqing Wang, Hao Wang,
  Yiya Wang, Yubo Wang, Jiajun Xu, Kexin Yang, Ruibin Yuan, Yuanhao Yue,
  Tianyang Zhan, Chun Zhang, Jinyang Zhang, Xiyue Zhang, Xingjian Zhang, Yue
  Zhang, Yongchi Zhao, Xiangyu Zheng, Chenghua Zhong, Yang Gao, Zhoujun Li,
  Dayiheng Liu, Qian Liu, Tianyu Liu, Shiwen Ni, Junran Peng, Yujia Qin, Wenbo
  Su, Guoyin Wang, Shi Wang, Jian Yang, Min Yang, Meng Cao, Xiang Yue,
  Zhaoxiang Zhang, Wangchunshu Zhou, Jiaheng Liu, Qunshu Lin, Wenhao Huang, and
  Ge~Zhang.
\newblock {SuperGPQA}: Scaling {LLM} evaluation across 285 graduate
  disciplines.
\newblock \emph{CoRR}, abs/2502.14739, 2025.
\newblock \doi{10.48550/ARXIV.2502.14739}.
\newblock URL \url{https://doi.org/10.48550/arXiv.2502.14739}.

\bibitem[Touvron et~al.(2023)Touvron, Martin, Stone, Albert, Almahairi, Babaei,
  Bashlykov, Batra, Bhargava, Bhosale, Bikel, Blecher, Canton{-}Ferrer, Chen,
  Cucurull, Esiobu, Fernandes, Fu, Fu, Fuller, Gao, Goswami, Goyal, Hartshorn,
  Hosseini, Hou, Inan, Kardas, Kerkez, Khabsa, Kloumann, Korenev, Koura,
  Lachaux, Lavril, Lee, Liskovich, Lu, Mao, Martinet, Mihaylov, Mishra,
  Molybog, Nie, Poulton, Reizenstein, Rungta, Saladi, Schelten, Silva, Smith,
  Subramanian, Tan, Tang, Taylor, Williams, Kuan, Xu, Yan, Zarov, Zhang, Fan,
  Kambadur, Narang, Rodriguez, Stojnic, Edunov, and
  Scialom]{DBLP:journals/corr/abs-2307-09288}
Hugo Touvron, Louis Martin, Kevin Stone, Peter Albert, Amjad Almahairi, Yasmine
  Babaei, Nikolay Bashlykov, Soumya Batra, Prajjwal Bhargava, Shruti Bhosale,
  Dan Bikel, Lukas Blecher, Cristian Canton{-}Ferrer, Moya Chen, Guillem
  Cucurull, David Esiobu, Jude Fernandes, Jeremy Fu, Wenyin Fu, Brian Fuller,
  Cynthia Gao, Vedanuj Goswami, Naman Goyal, Anthony Hartshorn, Saghar
  Hosseini, Rui Hou, Hakan Inan, Marcin Kardas, Viktor Kerkez, Madian Khabsa,
  Isabel Kloumann, Artem Korenev, Punit~Singh Koura, Marie{-}Anne Lachaux,
  Thibaut Lavril, Jenya Lee, Diana Liskovich, Yinghai Lu, Yuning Mao, Xavier
  Martinet, Todor Mihaylov, Pushkar Mishra, Igor Molybog, Yixin Nie, Andrew
  Poulton, Jeremy Reizenstein, Rashi Rungta, Kalyan Saladi, Alan Schelten, Ruan
  Silva, Eric~Michael Smith, Ranjan Subramanian, Xiaoqing~Ellen Tan, Binh Tang,
  Ross Taylor, Adina Williams, Jian~Xiang Kuan, Puxin Xu, Zheng Yan, Iliyan
  Zarov, Yuchen Zhang, Angela Fan, Melanie Kambadur, Sharan Narang,
  Aur{\'{e}}lien Rodriguez, Robert Stojnic, Sergey Edunov, and Thomas Scialom.
\newblock Llama 2: Open foundation and fine-tuned chat models.
\newblock \emph{CoRR}, abs/2307.09288, 2023.
\newblock \doi{10.48550/ARXIV.2307.09288}.
\newblock URL \url{https://doi.org/10.48550/arXiv.2307.09288}.

\bibitem[Wang et~al.(2025)Wang, Cheng, Peng, Bao, Li, Guo, Li, Zeng, Zhou, and
  Qiu]{DBLP:journals/corr/abs-2507-00018}
Bo~Wang, Qinyuan Cheng, Runyu Peng, Rong Bao, Peiji Li, Qipeng Guo, Linyang Li,
  Zhiyuan Zeng, Yunhua Zhou, and Xipeng Qiu.
\newblock Implicit reward as the bridge: {A} unified view of {SFT} and {DPO}
  connections.
\newblock \emph{CoRR}, abs/2507.00018, 2025.
\newblock \doi{10.48550/ARXIV.2507.00018}.
\newblock URL \url{https://doi.org/10.48550/arXiv.2507.00018}.

\bibitem[Wen et~al.(2025)Wen, Cai, Xiao, He, An, Duan, Du, Liu, Tanglifu, Lv,
  Zou, Deng, Jia, and Zhang]{wen-etal-2025-light}
Liang Wen, Yunke Cai, Fenrui Xiao, Xin He, Qi~An, Zhenyu Duan, Yimin Du,
  Junchen Liu, Tanglifu Tanglifu, Xiaowei Lv, Haosheng Zou, Yongchao Deng,
  Shousheng Jia, and Xiangzheng Zhang.
\newblock Light-{R}1: Curriculum {SFT}, {DPO} and {RL} for long {COT} from
  scratch and beyond.
\newblock In Georg Rehm and Yunyao Li, editors, \emph{Proceedings of the 63rd
  Annual Meeting of the Association for Computational Linguistics (Volume 6:
  Industry Track)}, pages 318--327, Vienna, Austria, July 2025. Association for
  Computational Linguistics.
\newblock ISBN 979-8-89176-288-6.
\newblock \doi{10.18653/v1/2025.acl-industry.24}.
\newblock URL \url{https://aclanthology.org/2025.acl-industry.24/}.

\bibitem[Wen et~al.(2024)Wen, Wang, and Sun]{wen2024mindmap}
Yilin Wen, Zifeng Wang, and Jimeng Sun.
\newblock {MindMap}: Knowledge graph prompting sparks graph of thoughts in
  large language models.
\newblock In \emph{62nd Annual Meeting of the Association for Computational
  Linguistics, ACL 2024}, pages 10370--10388. Association for Computational
  Linguistics (ACL), 2024.

\bibitem[Wolf et~al.(2020)Wolf, Debut, Sanh, Chaumond, Delangue, Moi, Cistac,
  Rault, Louf, Funtowicz, Davison, Shleifer, von Platen, Ma, Jernite, Plu, Xu,
  Scao, Gugger, Drame, Lhoest, and Rush]{DBLP:conf/emnlp/WolfDSCDMCRLFDS20}
Thomas Wolf, Lysandre Debut, Victor Sanh, Julien Chaumond, Clement Delangue,
  Anthony Moi, Pierric Cistac, Tim Rault, R{\'{e}}mi Louf, Morgan Funtowicz,
  Joe Davison, Sam Shleifer, Patrick von Platen, Clara Ma, Yacine Jernite,
  Julien Plu, Canwen Xu, Teven~Le Scao, Sylvain Gugger, Mariama Drame, Quentin
  Lhoest, and Alexander~M. Rush.
\newblock Transformers: State-of-the-art natural language processing.
\newblock In Qun Liu and David Schlangen, editors, \emph{Proceedings of the
  2020 Conference on Empirical Methods in Natural Language Processing: System
  Demonstrations, {EMNLP} 2020 - Demos, Online, November 16-20, 2020}, pages
  38--45. Association for Computational Linguistics, 2020.
\newblock \doi{10.18653/V1/2020.EMNLP-DEMOS.6}.
\newblock URL \url{https://doi.org/10.18653/v1/2020.emnlp-demos.6}.

\bibitem[Xie et~al.(2024)Xie, Aggarwal, and Ahmad]{xie-etal-2024-efficient}
Yong Xie, Karan Aggarwal, and Aitzaz Ahmad.
\newblock {Efficient Continual Pre-training for Building Domain Specific Large
  Language Models}.
\newblock In Lun-Wei Ku, Andre Martins, and Vivek Srikumar, editors,
  \emph{Findings of the Association for Computational Linguistics: ACL 2024},
  pages 10184--10201, Bangkok, Thailand, August 2024. Association for
  Computational Linguistics.
\newblock \doi{10.18653/v1/2024.findings-acl.606}.
\newblock URL \url{https://aclanthology.org/2024.findings-acl.606/}.

\bibitem[Yao et~al.(2023)Yao, Yu, Zhao, Shafran, Griffiths, Cao, and
  Narasimhan]{yao2023tree}
Shunyu Yao, Dian Yu, Jeffrey Zhao, Izhak Shafran, Tom Griffiths, Yuan Cao, and
  Karthik Narasimhan.
\newblock Tree of {T}houghts: Deliberate problem solving with large language
  models.
\newblock \emph{Advances in neural information processing systems},
  36:\penalty0 11809--11822, 2023.

\bibitem[Zhang and Soh(2024)]{zhang2024extract}
Bowen Zhang and Harold Soh.
\newblock {Extract, Define, Canonicalize}: An {LLM}-based framework for
  knowledge graph construction.
\newblock In \emph{Proceedings of the 2024 Conference on Empirical Methods in
  Natural Language Processing}, pages 9820--9836, 2024.

\bibitem[Zhu et~al.(2024)Zhu, Wang, Chen, Qiao, Ou, Yao, Deng, Chen, and
  Zhang]{zhu2024llms}
Yuqi Zhu, Xiaohan Wang, Jing Chen, Shuofei Qiao, Yixin Ou, Yunzhi Yao, Shumin
  Deng, Huajun Chen, and Ningyu Zhang.
\newblock {LLM}s for knowledge graph construction and reasoning: Recent
  capabilities and future opportunities.
\newblock \emph{World Wide Web}, 27\penalty0 (5):\penalty0 58, 2024.

\end{thebibliography}

\appendix

\section{Evaluation Benchmarks}
\label{section:appendix_benchmark}

Table \ref{table:evaluation_benchmarks} provides the details of our benchmarks. All our evaluation datasets and the 12K sampled U.S. cases for KG creation (Section \ref{section:evaluation_kg_generation}) are in English. All use of these artifacts are consistent with their intended use.
\begin{table*}[ht]
  \centering
  \begin{tabular}{llccc}
    \hline
    \multirow{2}{*}{\textbf{Benchmark}} &  \multirow{2}{*}{\textbf{Task Type}} & \textbf{Total} & \textbf{Sample} & \multirow{2}{*}{\textbf{Public}} \\
    & & \textbf{Tasks} & \textbf{Count} & \\ \hline
    \multirow{2}{*}{LexGLUE \citep{DBLP:journals/corr/abs-2304-12202}} & Classification and &  \multirow{2}{*}{6} & 1,000 &  \multirow{2}{*}{Yes} \\
     & MCQA & & for each task & \\
     \hline
    \multirow{2}{*}{LegalBench \citep{DBLP:conf/nips/GuhaNHRCKCPWRZT23}} & Yes/No, MCQA and &  \multirow{2}{*}{161} & 46 $\sim$ 10K &  \multirow{2}{*}{Yes} \\
    & Information Extraction & & per task & \\
    \hline
    \multirow{2}{*}{COLIEE \citep{DBLP:journals/rss/RabeloGKKYS22}} & Information Retrieval &  \multirow{2}{*}{4} & 500 $\sim$ 1.2K &  \multirow{2}{*}{Yes} \\
    & and Entailment & & per task & \\
    \hline
    SuperGPQA \citep{DBLP:journals/corr/abs-2502-14739} & MCQA & 1 & 656 & Yes \\
   \hline
    $\delta$-Stance \cite{DBLP:conf/acl/GuptaR025} & Classification & 2 & 14K & Yes \\
   \hline
  \end{tabular}
  \caption{\label{table:evaluation_benchmarks}
    Evaluation benchmarks. MCQA stands for Multiple Choice Question Answering. For SuperGPQA, we only evaluated on the \textit{Law} questions.
  }
\end{table*}

\newpage
\section{Grouping of Introductory Signals}
\label{section:delta_stance}

Table \ref{table:delta_stance} shows the grouping of all introductory signals into higher-level classes for our evaluation on the $\delta$-Stance dataset \cite{DBLP:conf/acl/GuptaR025}.
\begin{table}[ht]
  \centering
  \begin{tabular}{ll}
    \hline
    \textbf{Higher Level Class} & \multicolumn{1}{c}{\textbf{Signals}} \\ \hline
    directly supports & ``e.g.,'', ``accord'' \\ \hline
    indirectly supports & ``see'', ``see also'', ``cf.'' \\ \hline
    \multirow{2}{*}{contradicts} & ``but cf.'', ``but see'' \\
    & ``contra'' \\ \hline
    background & ``see generally'' \\ \hline
  \end{tabular}
  \caption{\label{table:delta_stance}
    Grouping of introductory signals for $\delta$-Stance. For the 3-class version in our evaluation, we removed the ``background'' class.
  }
\end{table}

\section{Model Training and Inference Details}
\label{section:hyperparameter}

We adopted Transformers \cite{DBLP:conf/emnlp/WolfDSCDMCRLFDS20}, DeepSpeed\footnote{https://github.com/deepspeedai/DeepSpeed}, Liger Kernel \cite{DBLP:journals/corr/abs-2410-10989} and Transformer Reinforcement Learning (TRL)\footnote{https://github.com/huggingface/trl} for training. Table \ref{table:hyperparameter} shows the detailed training hyperparameters. We did not conduct any parameter search; instead, we followed industry best practices to set the important parameters.

One ml.p4de.24xlarge instance has eight A100 GPUs, and each GPU has 80GB GPU RAM. For Llama-3.1-70B-Instruct, we employed three and four instances for SFT and DPO respectively; for Llama-3\_3-Nemotron-Super-49B-v1\_5, we utilized two and three instances for SFT and DPO respectively; for Qwen3-30B-A3B-Instruct-2507, we adopted one and two instances for SFT and DPO respectively.

In terms of training time, for Qwen3-30B-A3B-Instruct-2507, Llama-3\_3-Nemotron-Super-49B-v1\_5 and Llama-3.1-70B-Instruct, it needed 13, 2.5 and 3.5 hours for SFT respectively, and took 6, 2 and 2.5 hours for DPO respectively. For all model training, we trained for five epochs, and used our development set (Section \ref{section:evaluation_kg_generation}) for model selection. All our model training is fully reproducible. In total, our model training consumed 320 GPU hours.

We also employed SGLang\footnote{https://github.com/sgl-project/sglang} for inference; due to its non-reproducible nature, we ran each inference twice and reported the average. On our evaluation benchmarks, a single inference run can take up to 70 minutes (the COLIEE benchmark in Appendix \ref{section:appendix_benchmark}). We ran all inferences on a single instance. All our evaluations consumed a total of 480 GPU hours.

These libraries are open-source, and the base models are available from Hugging Face\footnote{https://huggingface.co/models}. The only proprietary artifact we use is Sonnet 3.5. All use of these artifacts are consistent with their intended use.

\begin{table}[ht]
  \centering
  \begin{tabular}{lcr}
    \hline
    \textbf{Hyperparameter} & \textbf{SFT} & \textbf{DPO} \\ \hline
    Learning rate & 1e-6 & 5e-7 \\
    Scheduler & \multicolumn{2}{c}{linear} \\
    Effective batch size & 384 & 192 \\
    Optimizer & \multicolumn{2}{c}{Adam} \\
    Attention & \multicolumn{2}{c}{FA2} \\
    Liger Kernel & Yes & No \\
    Model precision & \multicolumn{2}{c}{bfloat16} \\
    Gradient & \multicolumn{2}{c}{\multirow{2}{*}{Yes}} \\
    checkpointing & \\
    DeepSpeed & \multicolumn{2}{c}{Zero-3 without offload} \\
    Hardware & \multicolumn{2}{c}{ml.p4de.24xlarge} \\ 
    Training Data Size & 13.5K & 5.3K \\ \hline
  \end{tabular}
  \caption{\label{table:hyperparameter}
    Model training hyperparameters. SFT and DPO represent Supervised Fine-Tuning and Direct Preference Optimization respectively. \textit{Effective batch size} includes gradient accumulation.
  }
\end{table}

\newpage
\section{Llama-3.1-70B-Instruct vs. Llama-3.3-70B-Instruct}
\label{section:llama}

Table \ref{table:llama} compares the performance between Llama-3.1-70B-Instruct and Llama-3.3-70B-Instruct. First of all, on all benchmarks, Llama-3.1 had higher average performance than Llama-3.3, suggesting that it could be a stronger base model for training. Furthermore, Llama-3.1 also demonstrates better performance on 10 out of the 14 individual tasks, solidifying its advantage as a promising base model for further SFT and DPO. Finally, on the individual tasks where Llama-3.3 does achieve better performance than Llama-3.1, some of the differences are (extremely) small, such as SCOTUS, CaseHOLD and COLIEE Task-4. Based on the results, we conducted all our training by utilizing Llama-3.1-70B-Instruct as the base model.
\begin{table}[ht]
  \centering
  \begin{threeparttable}
  \begin{tabular}{llcc}
    \hline
    \textbf{Benchmark} & \textbf{Task} & \textbf{3.3} & \textbf{3.1} \\
    \hline
    \multirow{7}{*}{LexGLUE} & ECtHR A\tnote{$\bigstar$} & 69.2 & \underline{73.6} \\
    & ECtHR B\tnote{$\bigstar$} & 76.2 & \underline{80.0}  \\
    & EUR-LEX\tnote{$\bigstar$} & 31.7 & \underline{33.9}  \\
    & LEDGAR\tnote{$\bigstar$} & \underline{65.8} & 63.6  \\
    & SCOTUS\tnote{$\bigstar$} & \underline{67.7} & 67.5  \\
    & CaseHOLD\tnote{$\S$} & \underline{71.6} & \underline{71.6}  \\
    & Average & 63.7 & \textbf{65.0}  \\
   \hline
   \multirow{5}{*}{COLIEE} & Task-1\tnote{$\bigstar$} & \underline{32.6} & 30.8  \\
    & Task-2\tnote{$\bigstar$} & 49.2 & \underline{57.7}  \\
    & Task-3\tnote{$\triangle$} & 60.7 & \underline{61.3}  \\
    & Task-4\tnote{$\S$} & \underline{78.9} & 78.4  \\
    & Average & 55.4 & \textbf{57.0}  \\
   \hline
   \multirow{3}{*}{$\delta$-Stance\tnote{$\dag$}} & 3-class & 54.5 & \underline{55.1} \\
    & 4-class & 41.3 & \underline{42.6} \\
    & Average & 47.9 & \textbf{48.8} \\
   \hline
   \multicolumn{2}{l}{LegalBench\tnote{$\blacktriangle$}} & 76.7 & \textbf{78.0}  \\
   \hline
   \multicolumn{2}{l}{SuperGPQA (Law)\tnote{$\S$}} & 39.9 & \textbf{40.3}  \\
   \hline
  \end{tabular}
  \begin{tablenotes}
    \item[$\bigstar$]Micro-F1
    \item[$\dag$]Macro-F1
    \item[$\triangle$]Macro-F2
    \item[$\S$]Accuracy
    \item[$\blacktriangle$]Balanced Accuracy
  \end{tablenotes}
  \end{threeparttable}
  \caption{\label{table:llama}
    Comparison between the out-of-the-box version of Llama-3.3-70B-Instruct and Llama-3.1-70B-Instruct, denoted as 3.3 and 3.1 respectively.
  }
\end{table}

\newpage
\section{Prompts}
\label{section:appendix_prompt}

\subsection{IRAC KG Generation}
\label{section:appendix_prompt_kg}

Figure \ref{figure:prompt_kg_part1} and Figure \ref{figure:prompt_kg_part2} contain the prompt for producing the IRAC KG. Due to its length, we split it to two parts. We created the KGs using U.S. legal case opinions. Although these cases may contain personally identifying information, they are public data and are searchable via various online tools.
\begin{figure*}[htbp]
\begin{tcolorbox}[enhanced,colback=orange!5!white,colframe=orange!75!black,title=Prompt for Generating IRAC KG (Part I)]

\raggedright

You are tasked with building a knowledge graph by extracting entities and relations from a given legal case using the IRAC (Issue, Rule, Application, Conclusion) framework. This knowledge graph will be used for legal reasoning. Follow these instructions carefully to complete the task.\newline

First, carefully read and analyze the following legal case:\newline

<legal\_case>\newline
\{case\_opinion\}\newline
</legal\_case>\newline

Now, follow these steps to construct the knowledge graph:\newline

1. Entity Extraction:\newline
Extract the following types of entities from the legal case:\newline
- Case: The given case itself\newline
- CitedCase: Other cases cited in the given case\newline
- MaterialFact: Factual elements relevant to the case\newline
- LegalIssue: Legal issues presented in the given case\newline
- Conclusion: The court's ruling on a specific LegalIssue\newline
- Rule: Legal rules or principles applied in the case\newline
- Statute: Laws referenced in the case\newline
- Regulation: Regulations mentioned in the case\newline

For each entity, assign a unique ID, determine its type, and provide a succinct text representing the key aspects of the entity.\newline

2. Relation Extraction:\newline
Identify and extract the following relations between the entities:\newline
- CITES: From Case to CitedCase\newline
- REFERENCES: From Case to Statute or Regulation\newline
- ARISES\_FROM: From LegalIssue to MaterialFact\newline
- ADDRESSES: From Rule to LegalIssue\newline
- APPLIED\_TO: From Rule to MaterialFact\newline
- DERIVES\_FROM: From Rule to CitedCase, Statute, or Regulation\newline
- LEADS\_TO: From Rule to Conclusion\newline

For each relation, assign a unique ID, determine its type, and identify the source (from) and target (to) entities.\newline

\end{tcolorbox}
\caption{Prompt for generating IRAC KG (Part I).}
\label{figure:prompt_kg_part1}
\end{figure*}

\begin{figure*}[htbp]
\begin{tcolorbox}[enhanced,colback=orange!5!white,colframe=orange!75!black,title=Prompt for Generating IRAC KG (Part II)]

\raggedright

3. Output Formatting:\newline
Present your output as a JSON-formatted string that adheres to the following pydantic model structure. Ensure that all required fields (id\_, type\_, label\_ for entities; id\_, type\_, from\_, to\_ for relations) are properly filled.\newline

class Entity(BaseModel):\newline
    id\_: str = Field(description="entity ID")\newline
    type\_: str = Field(description="entity type")\newline
    label\_: str = Field(description="a succinct text representing the key aspects of the entity")\newline

class Relation(BaseModel):\newline
    id\_: str = Field(description="relation ID")\newline
    type\_: str = Field(description="relation type")\newline
    from\_: str = Field(description="source of relation")\newline
    to\_: str = Field(description="target of relation")\newline

class KG(BaseModel):\newline
    vertices\_: List[Entity] = Field(description="vertices from KG")\newline
    relations\_: List[Relation] = Field(description="relations from KG")\newline

4. Ensure that your output is comprehensive, capturing all relevant entities and relations from the given legal case. Be thorough in your analysis and precise in your entity and relation extractions.\newline

5. For entity type "CitedCase", only extract the most important cited cases that are relevant for resolving the legal issues.\newline

6. Do not generate any additional text, explanation or analysis.\newline

\end{tcolorbox}
\caption{Prompt for generating IRAC KG (Part II).}
\label{figure:prompt_kg_part2}
\end{figure*}

\subsection{Prompt for Generating the Explanation Section of the SFT Data}
\label{section:appendix_prompt_sft}

Figure \ref{figure:prompt_sft_part1} and Figure \ref{figure:prompt_sft_part2} show the prompt for producing the SFT training data. Due to its length, we split it to two parts.
\begin{figure*}[htbp]
\begin{tcolorbox}[enhanced,colback=orange!5!white,colframe=orange!75!black,title=Prompt for Generating the SFT Training Data (Part I)]

\raggedright

You are tasked with producing Instruction Finetuning (IFT) data for training a Large Language Model (LLM). The goal is to create a dataset that will teach the model how to identify legal issues and applicable rules based on case facts. This task is crucial for developing AI systems capable of assisting in legal analysis.\newline

First, you will be presented with the facts of a specific legal issue in the legal case. Read these carefully:\newline

<case\_facts>\newline
\{material\_facts\}\newline
</case\_facts>\newline

Analyze these facts thoroughly, considering potential legal issues that might arise from the situation described and potential legal rules that might be applicable to the facts and the legal issues.\newline

Now, here is one actual legal issue identified for this case, and the actual legal rules that are applicable to the given facts and the legal issue.\newline

<legal\_issue>\newline
\{legal\_issue\}\newline
</legal\_issue>\newline

<rules>\newline
\{rules\}\newline
</rules>\newline

Your task is to create IFT data that will teach an LLM to derive this issue from the given facts, and also to identify these legal rules that are applicable to the given case facts and the legal issue. To do this, follow these steps:\newline

1. Write a basic instruction for the LLM without including any details.\newline

2. In your response, include the above case facts, the legal issue, and the identified rules exactly as they were provided to you.\newline

3. Instruct the LLM to provide a brief explanation for the legal issue by relating it back to the given facts, and also on why the rules are applicable to the given case facts and the legal issue.\newline

\end{tcolorbox}
\caption{Prompt for generating the SFT training data (Part I).}
\label{figure:prompt_sft_part1}
\end{figure*}

\begin{figure*}[htbp]
\begin{tcolorbox}[enhanced,colback=orange!5!white,colframe=orange!75!black,title=Prompt for Generating the SFT Training Data (Part II)]

\raggedright

4. Format your output using the following XML format, and do not add any newlines or extra spaces between the XML elements:\newline
\begin{lstlisting}[language=XML, basicstyle=\ttfamily\small, breaklines=true]
<sft_data>
    <sft_input>
        <instruction>
            [Your instruction here]
        </instruction>
        <case_facts>
            [The list of case facts: 
            use separate <fact> tags for each case fact]
        </case_facts>
        <output_format>
            Provide your response in pretty print XML. Use top tag 
            legal_analysis, and the following sub-tags: 
            legal_issue: [The legal issue here]; rules: [A list of rules]; 
            explanation: [Brief explanation of the legal issue by relating 
            it to the case facts, and also on why the rules are applicable 
            to the given case facts and the legal issue].
        </output_format>
    </sft_input>
    <sft_output>
        <legal_issue>
            [The legal issue here]
        </legal_issue>
        <rules>
            [List each applicable rule using the tag rule]
        </rules>
        <explanation>
            [Brief explanation of the legal issue by relating it to the 
            case facts, and also on why the rules are applicable 
            to the given case facts and the legal issue]
        </explanation>
    </sft_output>
</sft_data>
\end{lstlisting}  

Now, create the IFT data for the case facts, the legal issue, and the rules provided earlier. Ensure your instruction is clear and comprehensive, and that your output follows the above XML format.\newline

Do not generate any additional text.\newline

\end{tcolorbox}
\caption{Prompt for generating the SFT training data (Part II).}
\label{figure:prompt_sft_part2}
\end{figure*}

\subsection{Prompt for the LLM Judge during DPO Data Generation}
\label{section:appendix_prompt_dpo}

Figure \ref{figure:prompt_dpo} presents the LLM Judge prompt for producing the DPO training data.
\begin{figure*}[htbp]
\begin{tcolorbox}[colback=orange!5!white,colframe=orange!75!black,title=LLM Judge Prompt for Generating the DPO Training Data]

\raggedright

You are a legal analyst tasked with determining whether specific legal rules can be applied to address a given legal issue.\newline

Your Task:\newline
Analyze whether the provided "Rules to Evaluate" are applicable to the stated legal issue, given the case facts and context.\newline

Information Provided:\newline
Case Facts (the relevant factual circumstances of the case):\newline
\{case\_facts\}\newline

Legal Issue (the specific legal question or dispute that needs to be resolved):\newline
\{legal\_issue\}\newline

Known Applicable Rules (a set of legal rules/statutes/precedents that are already established as relevant to this legal issue):\newline
\{chosen\_rules\}\newline

Rules to Evaluate (another set of legal rules that need to be assessed for applicability):\newline
\{rejected\_rules\}\newline

Your Analysis Should Address:\newline
- Relevance: Do the "Rules to Evaluate" address the same legal subject matter or related issues as the stated legal issue?\newline
- Factual Alignment: Are any factual elements required by these rules present in the given case facts?\newline
- Legal Compatibility: Are these rules consistent with or complementary to the "Known Applicable Rules"?\newline
- Jurisdictional Considerations: Are there any jurisdictional or procedural requirements that affect applicability?\newline
- Scope and Limitations: Are there any known boundaries to how these rules might apply?\newline

Provide your conclusion in JSON format. Use Rules as the top-level attribute; underneath it, include the evaluation for each rule in "Rules to Evaluate" with the following three attributes:\newline
- "Rule": the rule that is being assessed; use the rule text exactly as provided\newline
- "Applicability": Whether each rule in the "Rules to Evaluate" set is applicable (Yes/No/Potentially)\newline
- "Reasoning": Your reasoning for each determination\newline
\end{tcolorbox}
\caption{LLM Judge prompt for generating the DPO training data.}
\label{figure:prompt_dpo}
\end{figure*}

\end{document}